\documentclass[runningheads]{llncs}
\usepackage[T1]{fontenc}
\usepackage{amsmath, amsfonts, amssymb, mathtools}
\usepackage{url}

\usepackage{graphicx}
\usepackage{booktabs}
\newcommand{\vetx}{\boldsymbol{x}}
\newcommand{\vety}{\boldsymbol{y}}

\newcommand{\imI}{\mathbf{I}}
\newcommand{\imJ}{\mathbf{J}}

\newcommand{\bpm}{\begin{bmatrix}}
\newcommand{\epm}{\end{bmatrix}}

\usepackage{xcolor}

\newcommand{\new}[1]{#1}

\usepackage{hyperref}

\begin{document}
%
%\title{Learning a Reduced Mapping that Approaches a Condorcet Ordering for Vector-Valued Mathematical Morphology}
\title{Approximating Condorcet Ordering for Vector-valued Mathematical Morphology}
\titlerunning{Approximating Condorcet Ordering for Vector-Valued MM}
\author{Marcos Eduardo Valle\inst{1}\orcidID{0000-0003-4026-5110} \and
Santiago Velasco-Forero\inst{2}\orcidID{0000-0002-2438-1747}
\and
Joao Batista Florindo\inst{1}\orcidID{0000-0002-0071-0227}
\and
Gustavo (Jes\'us) Angulo\inst{3}%\orcidID{2222--3333-4444-5555}
}
\authorrunning{M.E. Valle et al.}
\institute{Universidade Estadual de Campinas (UNICAMP), Campinas, Brazil \and
Mines Paris - PSL University, Center for Mathematical Morphology (CMM), Fontainebleau, France \and Mines Paris - PSL University, Center for Applied Mathematics (CMA), Sophia-Antipolis, France
}
%\email{firstname.lastname@minesparis.psl.eu}
%
\maketitle              % typeset the header of the contribution
\begin{abstract}
Mathematical morphology provides a nonlinear framework for image and spatial data processing and analysis. Although there have been many successful applications of mathematical morphology to vector-valued images, such as color and hyperspectral images, there is still no consensus on the most suitable vector ordering for constructing morphological operators. This paper addresses this issue by examining a reduced ordering approximating the Condorcet ranking derived from a set of vector orderings. Inspired by voting problems, the Condorcet ordering ranks elements from most to least voted, with voters representing different orderings. In this paper, we develop a machine learning approach that learns a reduced ordering that approximates the Condorcet ordering. Preliminary computational experiments confirm the effectiveness of learning the reduced mapping to define vector-valued morphological operators for color images.   

%The abstract should briefly summarize the contents of the paper in 150--250 words.

\end{abstract}
\section{Introduction}

Mathematical morphology, a cornerstone of nonlinear image processing, presents unique challenges when applied to vector-valued data, such as color images or hyperspectral remote sensing datasets \cite{angulo07,aptoula07,velasco-forero14}. \new{Accordingly, morphological operators can be defined on complete lattices, which are ordered sets with well-defined extrema operations \cite{heijmans95,soille99}. However, there is no natural ordering for vectors, leading to the development of various approaches to vector-valued mathematical morphology. This paper applies concepts from social choice theory to address the challenge of selecting an appropriate ordering for vector-valued mathematical morphology. Social choice theory studies how individual preferences can be aggregated to reach a collective decision, such as electing a candidate through voting \cite{2016HandbookChoice}. In our context, we aim to establish consensus ordering for vector-valued mathematical morphology.} 

\new{
To our knowledge, integrating different orderings to define vector-valued mathematical morphology was first proposed by Lezoray \cite{Lezoray2021MathematicalOrderings}. Specifically, Lezoray proposed a weighted version of the Borda rule for combining stochastic permutation orderings. Although the Borda rule is computationally straightforward, it may produce an ordering that does not reflect the majority opinion \cite{Gehrlein2011VotingCoherence}. Instead of using the Borda rule, this paper considers Condorcet's principle, a more robust method for preference aggregation that prioritizes the most preferred candidates \cite{2016HandbookChoice,Lanctot2025SoftAgents}. The primary limitation of the Condorcet voting method is the computational complexity of the optimization techniques used to reach the consensus ordering.} By framing ordering relations as pairwise comparisons between vectors, we propose a learning-based framework to derive simplified representative mappings that approach Condorcet-optimal rankings. This paper is structured as follows: Section \ref{SecBasicMM} provides an overview of fundamental concepts in mathematical morphology, Section \ref{SecLearningCondorcet} introduces the methodology for learning Condorcet reduced orderings, Section \ref{SecComputationalExperiments} presents computational experiments, and Section \ref{sec:conclusions} concludes with final remarks and insights. 

\section{Basic Concepts on Mathematical Morphology}\label{SecBasicMM}

In this paper, we focus on morphological operators for vector-valued images. A vector-valued image, denoted as $\imI$, corresponds to a mapping $\imI: D \to \mathbb{V}$, where $D$ represents the image domain and $\mathbb{V} \subset \bar{\mathbb{R}}^d$, with $\bar{\mathbb{R}}=\mathbb{R} \cup \{-\infty,+\infty\}$ and $d \geq 2$, denotes a set of vector values. Throughout this paper, we also assume that the domain $D$ is a finite subset of a non-empty discrete additive Abelian group, $(\mathcal{E}, +)$.

% \subsection{Morphological Operators on Complete Lattices}

Mathematical morphology deals with non-linear operators successfully used for image processing and analysis \cite{heijmans95}. Generally speaking, morphological operators explore shapes and geometrical forms present in images. Complete lattices provide suitable frameworks for defining morphological operators \cite{ronse90}. A complete lattice $\mathbb{L}$ is a partially ordered set (poset) where every subset has both a supremum and an infimum \cite{birkhoff93}. The supremum and infimum of a set $X \subseteq \mathbb{L}$ are denoted by $\bigvee X$ and $\bigwedge X$, respectively. We assume that the vector value set $\mathbb{V}$ is a complete lattice in the following.

Dilations and erosions are two elementary morphological operations \cite{soille99}. For a given set \( S \subset \mathcal{E} \), referred to as a structuring element, the dilation and erosion of the image \( \mathbf{I} \) by \( S \) are defined by the equations below, respectively:  
\begin{equation} \label{eq:ero_dil}
\delta_S(\mathbf{I})(p) = \bigvee_{\substack{s \in S\\ p-s \in D}} \mathbf{I}(p-s) \quad \mbox{and} \quad 
\varepsilon_S(\mathbf{I})(p) = \bigwedge_{\substack{s \in S\\ p+s \in D}}\mathbf{I}(p+s), 
\quad \mbox{for all} \;\;p \in D.
\end{equation}
Other morphological operators are obtained by combining elementary morphological operators \cite{heijmans95}. For instance, the combinations of dilations and erosions yield opening and closing, which exhibit notable topological properties and function as non-linear image filters \cite{soille99}. Specifically, an opening is characterized by the composition of an erosion succeeded by a dilation, both using the same structuring element $S$. In contrast, a closing is defined as a dilation followed by an erosion. In mathematical terms, the opening $\gamma_{S}$ and the closing $\phi_{S}$ of an image $\imI$ by a structuring element $S$ are defined, respectively, by
\begin{equation} \label{eq:opening_closing}
\gamma_{S}(\mathbf{I}) = \delta_{S}\left(\varepsilon_{S}(\imI) \right) \quad \mbox{and} \quad 
\phi_{S}(\mathbf{I}) = \varepsilon_{S}\left(\delta_{S}(\imI) \right).
\end{equation}

We would like to highlight that dilations and erosions are defined in  \eqref{eq:ero_dil} through supremum and infimum operations. As a result, a partial order with well-defined extremal operations is sufficient for constructing morphological operators. However, unlike the one-dimensional (scalar) case, there is no inherent natural ordering for vectors, and various methods exist for ordering vector values. Examples of partial orderings used in vector-valued mathematical morphology include the marginal or Cartesian product ordering as well as conditional ordering schemes like the RGB-lexicographical ordering \cite{angulo07,aptoula07}. Moreover, despite not being partial orders but only pre-orders, reduced orderings have also been successfully used to develop vector-valued morphological operators \cite{goutsias95,velasco-forero14}. 

Several aspects must be considered when selecting a vector ordering for developing morphological operators in image processing tasks, including the \emph{``false color''} problem and the irregularity issue \cite{chevallier16,serra09}. The \emph{``false color''} problem occurs when a morphological operator introduces vector values that do not exist in the original image. This issue can be particularly problematic in applications such as satellite data analysis \cite{serra09}. One can avoid the \textit{``false color''} problem by ensuring that the partial order is total, that is, either $x \leq y$ or $y \leq x$ holds for any $x,y \in \mathbb{L}$.

With a total ordering, the extreme operators yield elements of the original image, preventing the creation of new vector values. Formally, for any finite set \(X\), we have \(\bigvee X \in X\) and \(\bigwedge X \in X\) when the supremum and infimum are calculated using a total ordering. However, one drawback of morphological operators based on total orders is that they may introduce irregularities and aliasing in the processed images \cite{chevallier16}. While these irregularities can negatively impact specific tasks, such as border detection, this paper focuses on morphological operators defined through total orders. Specifically, our approach mitigates the \textit{``false color''} problem by utilizing a reduced ordering combined with look-up tables (LUTs) \cite{velasco-forero14}. 

A reduced ordering is defined by a surjective mapping $h: \mathbb{V} \to \mathbb{L}$, which maps the value set $\mathbb{V}$ to a complete lattice $\mathbb{L}$. The partial order $\leq_{\mathbb{L}}$ on $\mathbb{L}$ induces a pre-order on $\mathbb{V}$ as follows:
\begin{equation} \label{eq:reduced}
\vetx \leq_h \vety \iff h(\vetx) \leq_\mathbb{L} h(\vety).
\end{equation}
In this context, the mapping \( h \) is called a reduced mapping or simply an \( h \)-mapping. It is important to note that a reduced ordering is not a partial order but rather a pre-order, as it may lack antisymmetry. This means that the inequalities \( \vetx \leq_h \vety \) and \( \vety \leq_h \vetx \) do not necessarily imply that \( \vetx = \vety \). Despite this limitation, reduced orderings can be effectively utilized to develop vector-valued morphological operators. By combining \eqref{eq:reduced} with LUTs, one can define computationally efficient vector-valued morphological operators. Furthermore, alongside a LUT, a reduced ordering yields a total ordering on the image's finite set of vector values. \new{We would like to point out that the use of reduced ordering and LUT for defining morphological operators is detailed in \cite{velasco-forero14}, where Algorithm 1 illustrates the implementation of a vector-valued erosion using MATLAB-like notation.} 

We would like to conclude this section by recalling that the surjective mapping \( h: \mathbb{V} \to \mathbb{L} \) can be established using various machine learning techniques, including both supervised and unsupervised methods. In the approach detailed in the following sections, the $h$-mapping is represented by a neural network trained on an image dataset to generate a consensus ordering.

\section{Learning Condorcet Reduced Orderings}\label{SecLearningCondorcet}

In practical scenarios, selecting a reduced mapping for a specific morphological image processing task can be somewhat arbitrary. In this paper, we address the challenge of identifying an appropriate reduced mapping $h$ by examining a family \( \mathcal{H} = \{ h_1, \ldots, h_n \} \) of possible \( h \)-mappings, called a \textit{profile} in social choice literature \cite{Zwicker2016IntroductionVoting}. Rather than using the common majority voting approach found in ensemble methods \cite{kuncheva14}, our approach seeks to establish an $h$-mapping that approaches a consensus through the orderings derived from the reduced mappings in \( \mathcal{H} \). Such a consensus of orderings is known as the Condorcet ordering in voting methods.

\subsection{Condorcet Ordering}

The concept of a Condorcet order is inspired by the contributions of \emph{Marie Jean Antonie Nicolas de Caritat} (1743-1794), known as the \emph{Marquis de Condorcet}, on the social choice field during the period of the Enlightenment. A social choice method aims to elect a candidate from a set of voters. Briefly, a candidate is considered a \emph{Condorcet winner} if it is the majority-preferred option in pairwise comparisons \cite{Zwicker2016IntroductionVoting}. However, we would like to remark that the Condorcet winner is not always well-defined. \new{For instance, the Condorcet voting paradox shows that there is no Condorcet winner when the pairwise majority rule yields a non-transitive relation.} 

In this paper, we speak of a Condorcet order $\leq^*$ as a total order on \(\mathbb{V}\) such that, if a subset \(\mathbf{X} \subseteq \mathbb{V}\) has a Condorcet winner, that winner equals \(\bigvee^* \mathbf{X}\). Furthermore, the problem of finding a Condorcet ordering is based on the \emph{Kemeny-Young method}, a Condorcet-consistent approach that yields the Condorcet winner when it exists \cite{2016HandbookChoice,Young1988CondorcetsVoting}. Accordingly, this method selects a total order $\leq^*$ that minimizes the Kemeny score over a family $\mathcal{G} = \{\leq_1,\leq_2,\ldots,\leq_m\}$ of $m$ total orderings \cite{Young1978APrinciple}. The following reviews a combinatorial linear optimization problem detailed by Marcotorchino and Michaud for finding a Condorcet ordering \cite{Marcotorchino1982AgregationAutomatique}.

First of all, recall that a total order $\leq$ on $\mathbb{V}$ is a binary relation that is reflexive, transitive, antisymmetric, and strongly connected. A binary relation is strongly connected if either $\vetx \leq \vety$ or $\vety \leq \vetx$ holds true for all $\vetx,\vety \in \mathbb{V}$. We can identify the total order on a finite set $\mathbb{V}_n = \{\vetx_1,\ldots,\vetx_n\}$ with a matrix $R \in \{0,1\}^{n \times n}$ such that
$r_{ij} = [\vetx_i \leq \vetx_j]$, for all $i,j=1,\ldots,n$, where 
\begin{equation}
    \label{eq:indicator}
    [\vetx \leq \vety] =  \begin{cases}
    1, & \vetx \leq \vety,\\
    0, & \text{otherwise},
    \end{cases}
\end{equation}
denotes the indicator function of the binary relation. Furthermore, $R$ is the matrix associated with a total order if, and only if, $r_{ii}=1$ for all $i=1,\ldots,n$, and the following holds for distinct indexes $i,j,k$ (i.e., $i \neq j$, $j \neq k$, and $k \neq i$):
\begin{equation} 
\label{eq:linear_restrictions}
r_{ij}+r_{jk}-r_{ik} \leq 1 \quad \text{and} \quad r_{ij}+r_{ji}=1.
\end{equation} 
We would like to note that \( r_{ii}=1 \) ensures reflexivity. The first inequality in \eqref{eq:linear_restrictions} guarantees the transitivity property. Finally, the second identity in \eqref{eq:linear_restrictions} addresses the antisymmetry and strong connectedness properties of a total order \cite{Marcotorchino1982AgregationAutomatique}.

Given a family $\mathcal{G} = \{\leq_1,\leq_2,\ldots,\leq_m\}$ of total orderings on a finite set $\mathbb{V}_n = \{\vetx_1,\ldots,\vetx_n\}$, the average vote margin of $\vetx$ under $\vety$ is given by
\begin{equation}
    \label{eq:delta}
    \delta(\vetx,\vety) = \frac{1}{m} \sum_{k=1}^m \big( [\vetx \leq_k \vety] - [\vety \leq_k \vetx] \big), \quad \forall \vetx,\vety \in \mathbb{V}.
\end{equation}
Note that $\delta(\vetx, \vety)$ calculates the average difference in the less than or equal count between $\vetx$ and $\vety$. Therefore, $\delta(\vetx,\vety) > 0$ if and only if $\vetx$ is less than or equal to $\vety$ in most orderings in $\mathcal{G}$. 

A Condorcet order $\leq^*$ derived from $\mathcal{G}$ corresponds to the total ordering on $\mathbb{V}_n=\{\vetx_1,\dots,\vetx_n\}$ associated with a matrix $R^* \in \mathbb{R}^{n \times n}$ that solves the following combinatorial optimization problem\footnote{We would like to point out that we obtain a minimization problem by considering a ``less than or equal'' relation in \eqref{eq:delta}. Dually, a ``greater than or equal'' relation yields a maximization problem. We considered a minimization problem because deep learning libraries often minimize the loss.} where $\delta_{ij} \equiv \delta(\vetx_i,\vetx_j)$:
\begin{equation}
    \label{eq:CondorcetProblem}
    \left\{ \begin{array}{rl}
    R^* = \displaystyle{\mathop{\text{argmin}}_{R = (r_{ij})}} & \displaystyle{\sum_{i=1}^n \sum_{j=1}^n \delta_{ij} r_{ij}},\\
    \text{subject to } & r_{ij}+r_{ij} = 1, \quad \forall i \neq j,\\
    & r_{ij}+r_{jk}-r_{ik} \leq 1, \quad \forall i \neq j, j \neq k, k \neq i,\\
    & r_{ij} \in \{0,1\}.
    \end{array} \right.
\end{equation}
Unfortunately, the problem described in \eqref{eq:CondorcetProblem} is NP-hard, making its computational complexity prohibitive for large $n$.
An alternative approach for approximating the solution to \eqref{eq:CondorcetProblem} is presented in the following subsection, based on \cite{Lanctot2025SoftAgents}.

\subsection{Soft Condorcet Ordering}

In a recent study, Lanctot et al. proposed a ranking method that approximates the Kemeny-Young voting system using score ratings and the logistic function \cite{Lanctot2025SoftAgents}. The following briefly reviews the key steps in their approach to determining the Condorcet ordering.

First, from lattice theory, we know that any finite totally ordered set is isomorphic to $\{1,\ldots,n\}$ \cite{birkhoff93}. By transitivity, a finite set \(\mathbb{V}_n = \{\vetx_1, \ldots, \vetx_n\}\) endowed with a total order can be associated with a corresponding set of scores \(\mathcal{S}_n = \{s_1, \ldots, s_n\} \subset \mathbb{R}\) with the natural order \(\leq_{\mathbb{R}}\) of real numbers. Formally, we have \(\vetx_i \leq \vetx_j\) if and only if \(s_i \leq_{\mathbb{R}} s_j\). Thus, determining a total order for $\mathbb{V}_n$ reduces to finding an appropriate score set $\mathcal{S}_n$. In particular, the problem \eqref{eq:CondorcetProblem} for finding the Condorcet ordering $\leq^*$ using the Kemeny-Young method becomes
\begin{equation}
    \label{eq:HardCondorcetOptimization}
    \mathcal{S}_n^* = \mathop{\text{argmin}}_{\mathcal{S}_n=\{s_1,\ldots,s_n\}}  \displaystyle{\sum_{i=1}^n \sum_{j=1}^n \delta_{ij} [s_i \leq_{\mathbb{R}} s_j]},
\end{equation}
A solution $R^*$ to \eqref{eq:CondorcetProblem} is derived from $\mathcal{S}_n^*$ by setting $r^*_{ij} = [s_i^* \leq_{\mathbb{R}} s_j^*]$. In this case, the constraints in \eqref{eq:CondorcetProblem} follow directly from assuming that $\mathcal{S}_n^*$ is equipped with the natural ordering of real numbers.

On the downside, \eqref{eq:HardCondorcetOptimization} relies on a piecewise constant step function with a discontinuity at $s_i=s_j$, making it extremely difficult to solve. To address this issue, Lanctot et al. replace the step function $[s_i \leq s_j]$ with a smooth approximation using a scaled logistic function:
\begin{equation}
    \sigma(s_j-s_i) = \frac{1}{1+e^{(s_j-s_i)/\tau}},
\end{equation}
where $\tau>0$ is the steepness parameter, controlling the transition speed between the asymptotic values $0$ and $1$. This substitution transforms the hard optimization problem in \eqref{eq:HardCondorcetOptimization} into the soft Condorcet optimization (SCO) problem
\begin{equation}
    \label{eq:SoftCondorcetOptimization}
    \mathcal{S}_n^* = \mathop{\text{argmin}}_{\mathcal{S}_n}  \displaystyle{\sum_{i=1}^n \sum_{j=1}^n \delta_{ij} \sigma(s_j-s_i)},
\end{equation}
which can be solved via gradient descent or sigmoidal programming \cite{Lanctot2025SoftAgents}. Empirical results indicate that both methods yield similar outcomes, with sigmoidal programming being recommended only for relatively small $n$ due to computational demands. Thus, we adopt a gradient descent-based approach in this paper.

Summarizing, the soft Condorcet optimization problem proposed by Lanctot et al. is applied as follows: Given a family \(\mathcal{G} = \{\leq_1, \leq_2, \ldots, \leq_m\}\) of \(m\) total orders on a finite set \(\mathbb{V}_n = \{\vetx_1, \ldots, \vetx_n\}\), a consensus total order \(\leq^*\) is generated by establishing that \(\vetx_i \leq^* \vetx_j\) if and only if \(s_i^* \leq_{\mathbb{R}} s_j^*\). Here, \(\mathcal{S}_n^* = \{s_1^*, \ldots, s_n^*\}\) represents a scoring set that solves the optimization problem \eqref{eq:SoftCondorcetOptimization}. \new{It is important to note that the optimal score set \(\mathcal{S}_n^*\), and consequently the consensus total order \(\leq^*\), depends on both \(\mathcal{G}\) and the finite subset \(\mathbb{V}_n \subseteq \mathbb{V}\)}. Specifically, if we were to replace \(\mathbb{V}_n\) with a different set \(\mathbb{V}_{n'}'\), the soft optimization problem would need to be solved again. \new{Therefore, the consensus total order \(\leq^*\) cannot be generalized across different finite subsets of the value set. Using \(h\)-orderings, we present a new approach to approximate the Condorcet consensus in the following section. Learning a reduced ordering enables the consideration of subsets (batches) of $\mathbb{V}_n$ as well as the ability to generalize to new values.}

\subsection{Learning from Reduced Orderings} \label{ssec:learning_Condorcet}

This paper primarily focuses on identifying a reduced mapping \( h^*:\mathbb{V} \to \mathbb{R} \), refered to as a Condorcet \( h^* \)-mapping, that captures the consensus of a family \( \mathcal{H} = \{h_1,\ldots,h_m\} \) of \( h \)-mappings. The Condorcet \( h^* \)-mapping is defined by a machine learning model that learns its parameters from a training set denoted as \(\mathbb{V}_n = \{\vetx_1, \ldots, \vetx_n\}\). For instance, \(\mathbb{V}_n\) can represent the set of colors extracted from a single image or the combined set of colors from multiple images. The finite set $\mathbb{V}_n$ can be considered a finite subset of the larger value set \(\mathbb{V}\). 

Specifically, let \( h:\Theta \to \mathbb{R}^\mathbb{V} \) represent a parameterized regressor, where \( h(\theta):\mathbb{V} \to \mathbb{R} \) is an \( h \)-mapping for each parameter vector \( \theta \in \Theta \), with \( \Theta \) being the set of all possible parameter vectors. The optimal parameter vector \( \theta^* \) for the machine learning model is determined by minimizing the soft Kemeny-Young loss function, as introduced by Lanctot et al., across both \( \mathbb{V}_n \) and \( \mathcal{H} \). Accordingly, based on \eqref{eq:delta}, we define the average voting margin of $\vetx_i$ under $\vetx_j$ by means of the following equation for all $i,j=1,\ldots,n$:
\begin{equation}
    \label{eq:delta_k}
    \delta_{ij} = \frac{1}{m} \sum_{k=1}^m 
    [h_k(\vetx_i) \leq_{\mathbb{R}} h_k(\vetx_j)]
    - [h_k(\vetx_j) \leq_{\mathbb{R}} h_k(\vetx_i)].
\end{equation}
Note that, instead of considering a family of total orderings, the quantity $\delta_{ij}$ is calculated over the family of reduced mappings $\mathcal{H} = \{h_1,\ldots,h_m\}$ using the natural order of real numbers. 
The soft Kemeny-Young loss function used for training the reduced mapping over $\mathcal{H}$ and $\mathbb{V}_n$ is defined by the following equation as a function of the regressor parameters $\theta \in \Theta$: 
\begin{equation}
    \label{eq:soft-loss}
    \mathcal{L}(\mathcal{H},\mathbb{V}_n)(\theta) = \sum_{i=1}^{n} \sum_{j=1}^{n} \delta_{ij} \sigma\big(h(\theta)(\vetx_j)-h(\theta)(\vetx_i)\big).
\end{equation}
% We would like to remark that, although we typically work with images of the same size, the loss function \eqref{eq:soft-loss} is well-defined for a training set $\mathbb{V}_n$ composed of the values of images of varying widths and lengths. The only requirement is that all the images must share the same set of values $\mathbb{V}$.

Concluding, given a family of $h$-mappings $\mathcal{H} =\{h_1,\ldots,h_m\}$ and a finite set of training values $\mathbb{V}_n=\{\vetx_1,\ldots,\vetx_n\} \subset \mathbb{V}$, the Condorcet $h^*$-mapping is given by $h^* \equiv h(\theta^*):\mathbb{V} \to \mathbb{R}$, where $\theta^*$ solves the optimization problem
\begin{equation}
    \theta^* = \mathop{\text{argmin}}_{\theta \in \Theta} \mathcal{L}(\mathcal{H},\mathbb{V}_n)(\theta).
\end{equation}
Finished the training phase, the Condorcet $h^*$-mapping can be utilized to define morphological operators, which serve as a consensus of the morphological operators derived from the reduced mappings $h_1, \ldots, h_m$. Additionally, due to the model's generalization capability, the Condorcet $h^*$-mapping can process or analyze new images not included in the training set. The computational experiments reported in the following section illustrate these remarks.

\subsection{Borda Rule}
\label{sec:Borda_Rule}

\new{
The Borda rule, proposed by \textit{Jean-Charles de Borda} (1733–1799) in 1770 during the Enlightenment period, serves as an alternative method to the Condorcet consensus for making social choices \cite{2016HandbookChoice}. Unlike the Condorcet consensus, which uses pairwise comparisons to establish a binary relation that results in a total order, the Borda rule ranks candidates by aggregating scores that reflect voters' preferences. The following briefly reviews the Borda rule.}

\new{
Consider a finite set \(\mathbb{V}_n = \{\vetx_1,\ldots,\vetx_n\} \subseteq \mathbb{V}\) (set of candidates) and let \(\mathcal{G} = \{\leq_1, \leq_2, \ldots, \leq_m\}\) be a family of \(m\) total orderings on \(\mathbb{V}_n\), reflecting the preferences of voters. The Borda score of a candidate $\vetx_i \in \mathbb{V}_n$ is determined by averaging the ranks of $\vetx_i$ across the ordered lists of all voters' preferences. Using the indicator function defined by \eqref{eq:indicator}, the Borda score of $\vetx_i$ is defined by
\begin{equation}
\label{eq:Borda_score}
\mathcal{B}(\vetx_i) = \frac{1}{m (n-1)}\sum_{k=1}^m \sum_{j \neq i}^n [\vetx_j \leq_k\vetx_i], \quad \forall i=1,\ldots,n.
\end{equation}
Note that \(\mathcal{B}(\vetx_i) \in [0, 1]\), where $\mathcal{B}(\vetx_i) = 0$ and $\mathcal{B}(\vetx_i) = 1$ signify that $\vetx_i$ is the least and the most preferred choice for all voters, respectively. The Borda rule establishes a total ordering \(\mathbb{V}_n\), denoted as ``\(\leq_B\)'', such that \(\vetx_i \leq_B \vetx_j\) when \(\mathcal{B}(\vetx_i) \leq_{\mathbb{R}} \mathcal{B}(\vetx_j)\). Like reduced orderings, the total order ``$\leq_B$'' can be computed by combining the Borda scores with a look-up table, as detailed in \cite{velasco-forero14}.
}

\new{The Borda rule is computationally cheaper than our proposed approach, which learns a reduced mapping that approximates the Condorcet consensus. Accordingly, instead of solving the optimization problem \eqref{eq:SoftCondorcetOptimization}, the Borda rule requires the computation of the scores by means of \eqref{eq:Borda_score} and a sorting operation. On the downside, the Borda rule can fail to elect a candidate who is top-ranked by the majority of the voters \cite{2016HandbookChoice}. For example, consider a set $\mathbb{V}_3 = \{\vetx_1,\vetx_2,\vetx_3\}$ with three candidates and let $\mathcal{G} = \{\leq_1,\ldots,\leq_5\}$ be a family of total orderings representing the preferences of five voters. Suppose the voter's preference are $\vetx_1 \leq_i \vetx_2 \leq_i \vetx_3$ for $i=1,2,3$ and $\vetx_3\leq_i \vetx_1 \leq_i \vetx_2$ for $i=4,5$. Note that $\vetx_3$ is the Condorcet winner because it is the preferred candidate of the majority of the voters. Moreover, following the majority of the voters, the Condorcet consensus yields the total order $\vetx_1 \leq^* \vetx_2 \leq^* \vetx_3$. In contrast, the Borda scores are $\mathcal{B}(\vetx_1)=2/10$, $\mathcal{B}(\vetx_2)=7/10$, and $\mathcal{B}(\vetx_1)=6/10$. As a consequence, the Borda rule yields $\vetx_1 \leq_B \leq \vetx_3 \leq_B \vetx_2$. Note that $\vetx_2$ is the largest element (the elected candidate) despite not being the Condorcet winner (the candidate preferred by the majority). 
}

\new{In conclusion, although the Borda rule is computationally less expensive than the Condorcet method, it may not adequately reflect the majority preference in social choices. In the field of mathematical morphology, Lezoray has successfully applied the Borda rule to define morphological operators that are generated using stochastic permutation orderings \cite{Lezoray2021MathematicalOrderings}. For this reason, we include the Borda rule for comparison purposes in the following section.}

\section{Computational Experiments}\label{SecComputationalExperiments}

Let us illustrate the approach described in the previous section for learning an $h$-mapping that closely approximates the Condorcet ordering derived from a family of reduced mappings $\mathcal{H} = \{h_1, \ldots, h_m\}$ and a set of RGB colors $\mathbb{V}_n$. \new{For comparison purposes, we also consider the Borda rule described in Section \ref{sec:Borda_Rule}.}
The Jupyter Notebook that outlines the computational experiments is available at \href{https://github.com/mevalle/CondorcetOrdering}{github.com/mevalle/CondorcetOrdering}.

\begin{figure}[t]
    \centering
    \begin{tabular}{cccccc}
    $\imI$ & $\gamma_S^{h_1}(\imI)$ & $\gamma_S^{h_2}(\imI)$ & $\gamma_S^{h_3}(\imI)$ & $\gamma_S^{B}(\imI)$ & $\gamma_S^{h^*}(\imI)$ \\ 
    \includegraphics[width=0.16\linewidth]{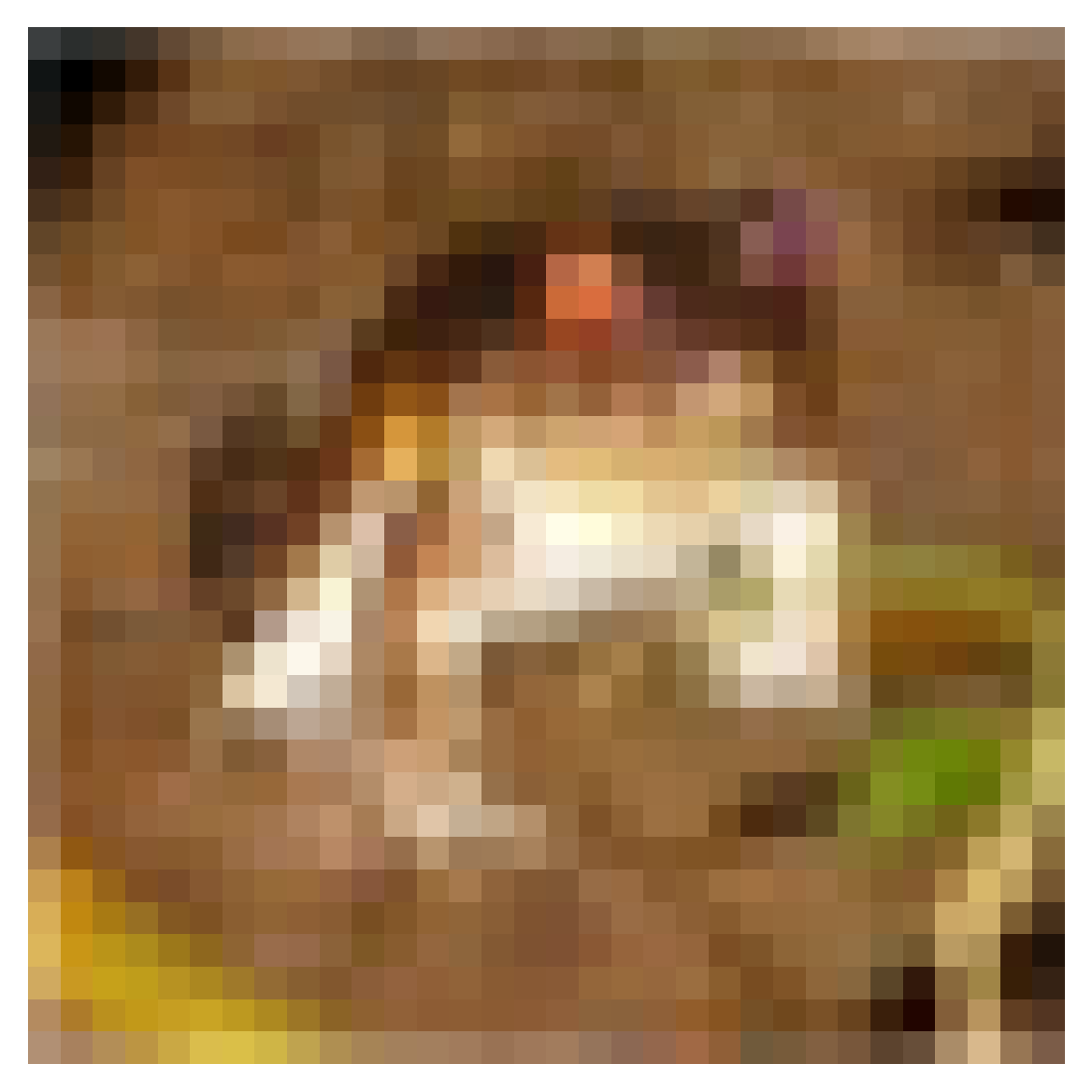} & 
    \includegraphics[width=0.16\linewidth]{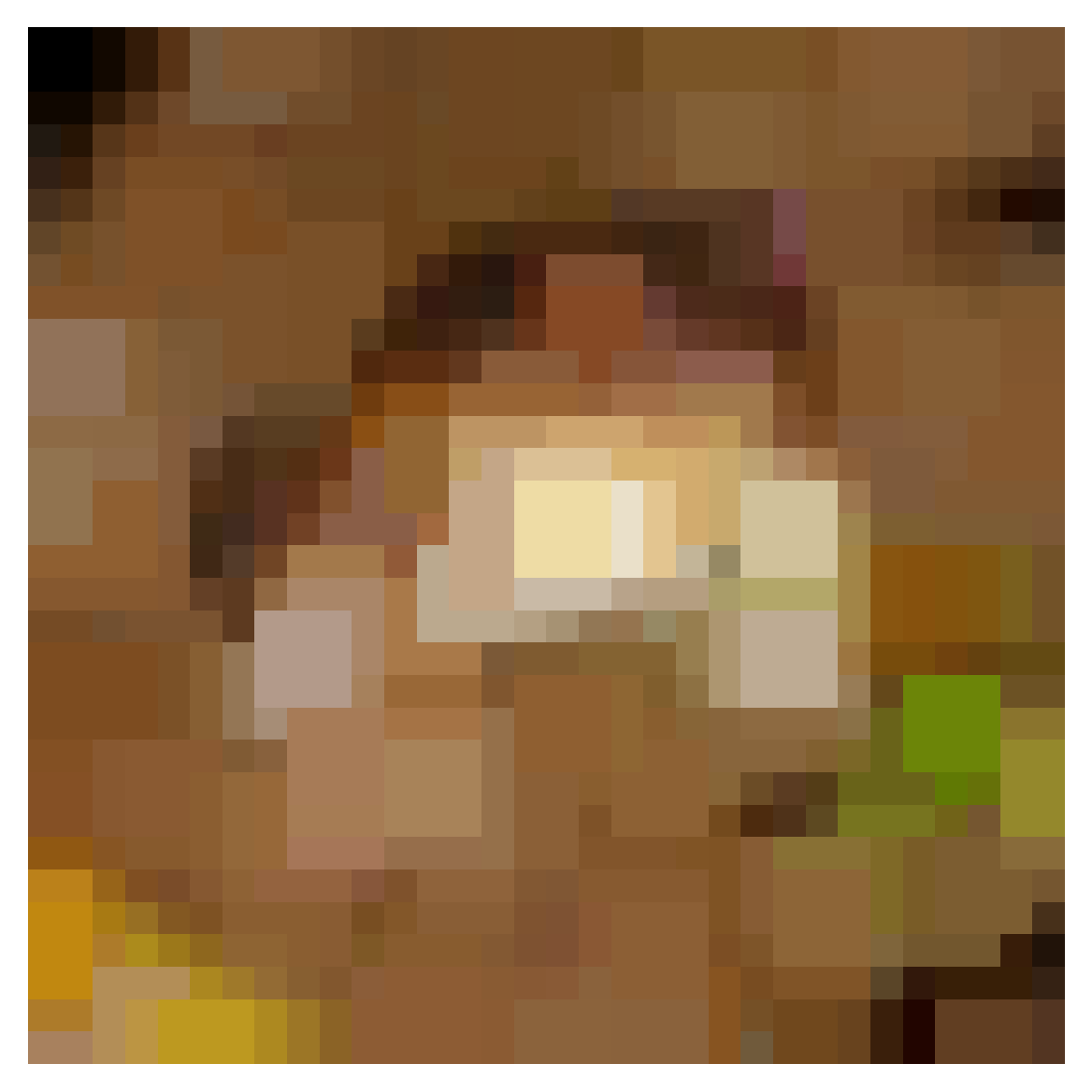} &
    \includegraphics[width=0.16\linewidth]{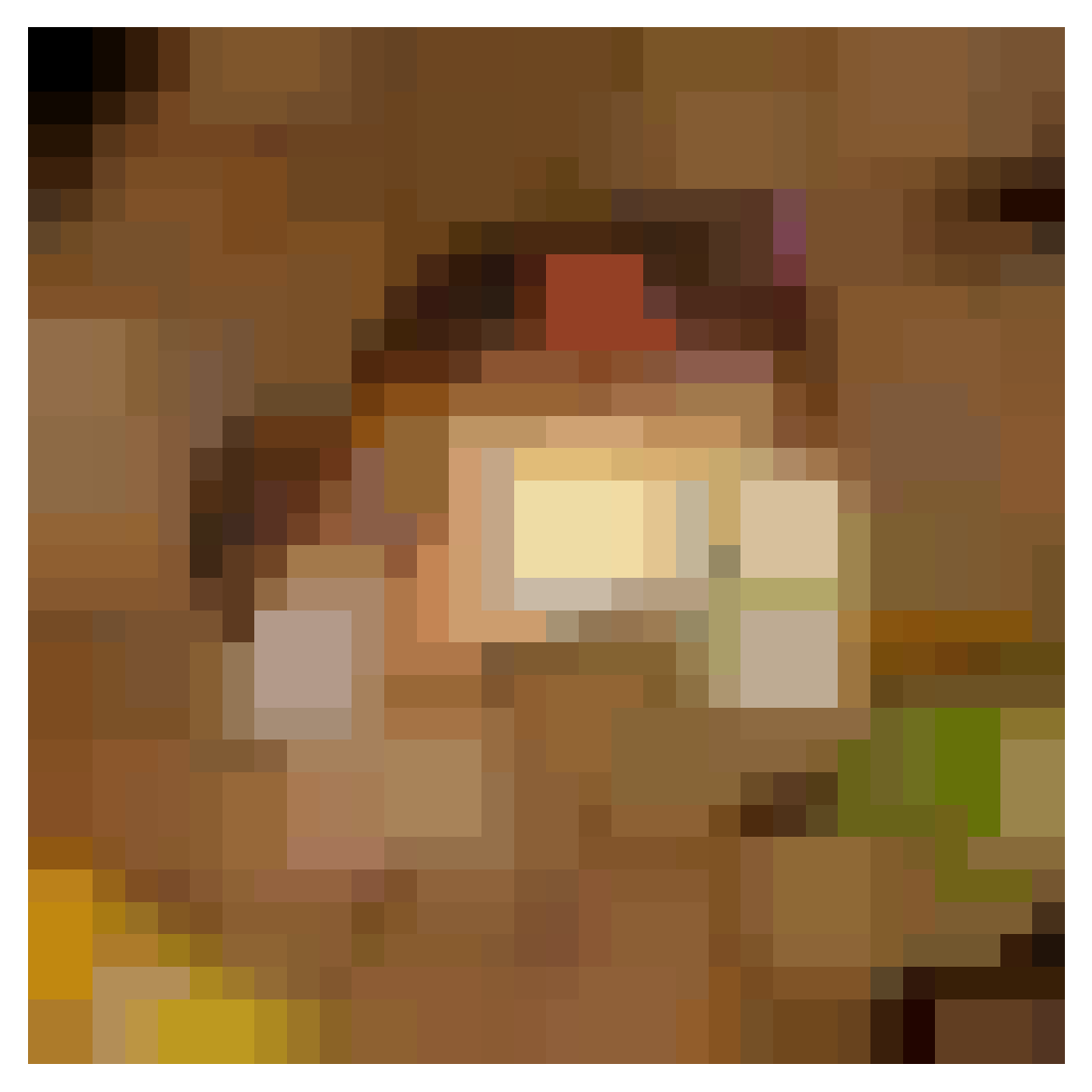} &
    \includegraphics[width=0.16\linewidth]{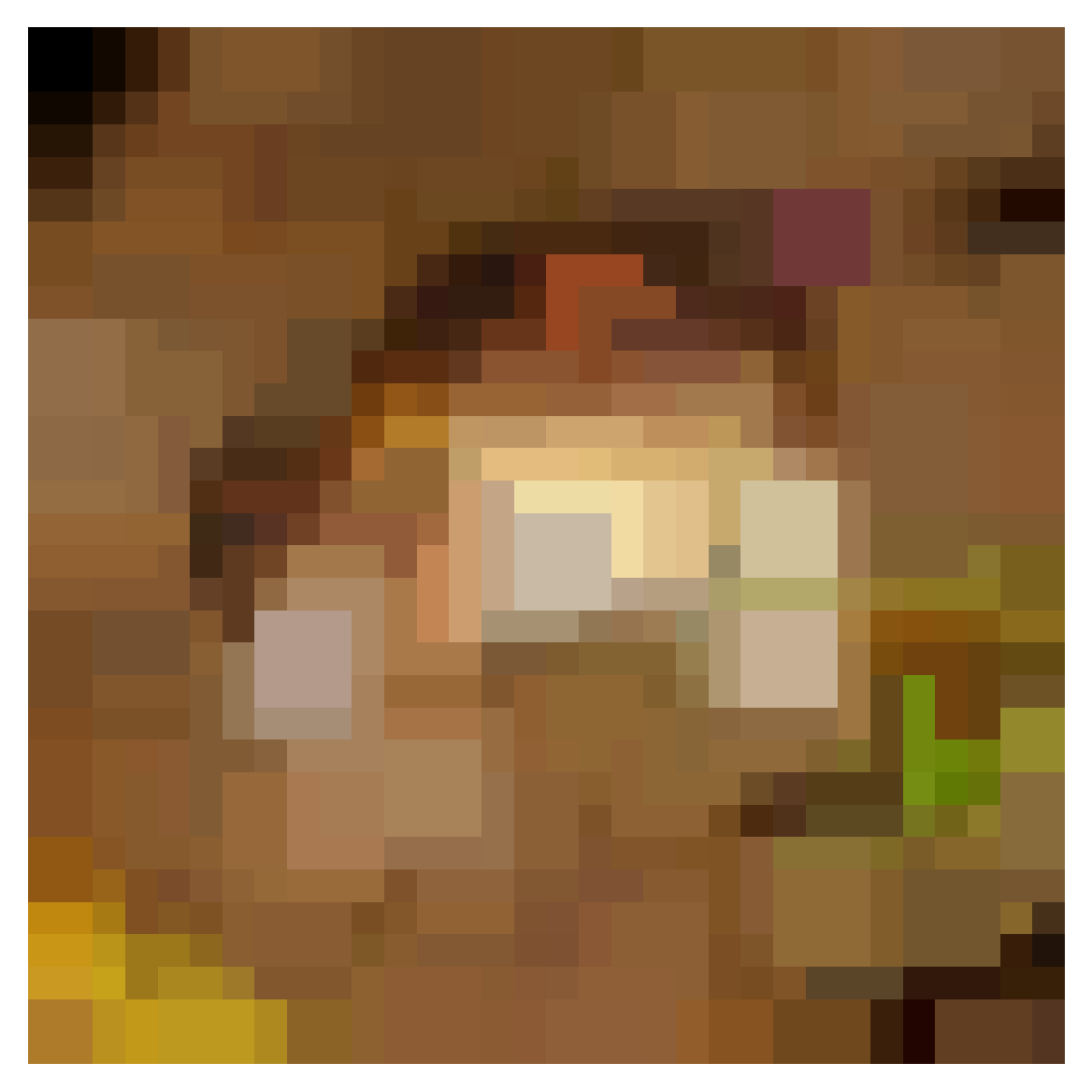} &
    \includegraphics[width=0.16\linewidth]{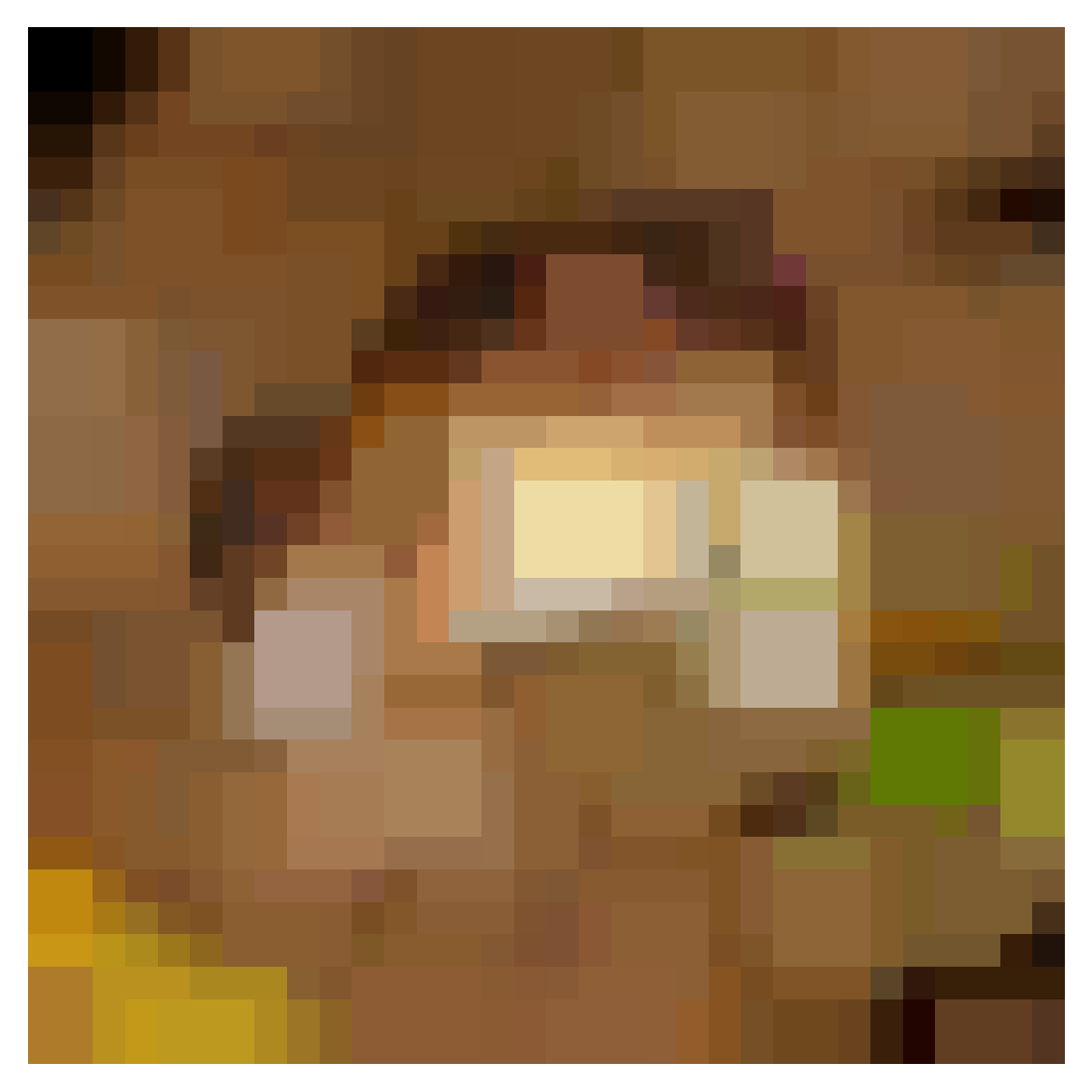} &
    \includegraphics[width=0.16\linewidth]{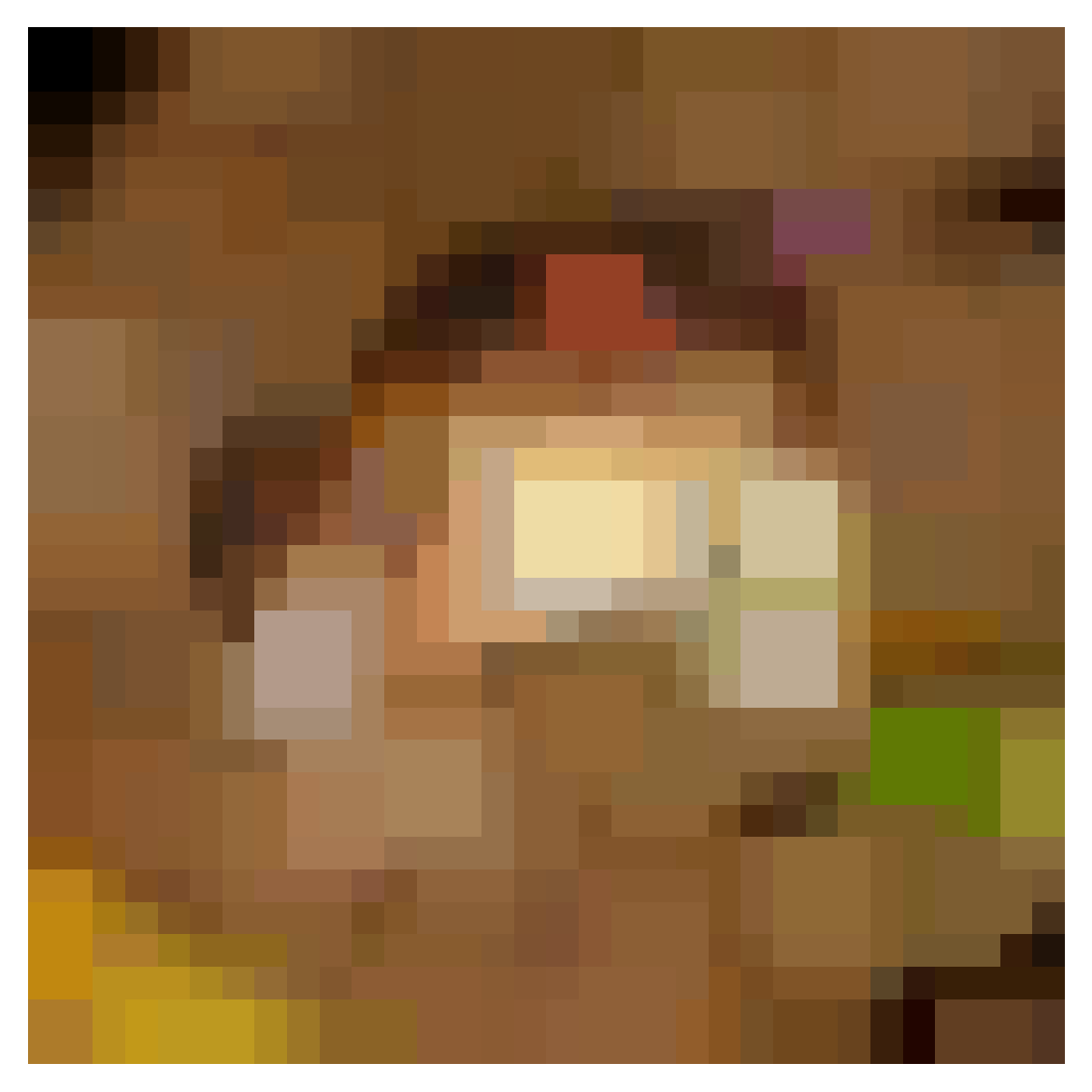} 
    \\ 
    \includegraphics[width=0.16\linewidth]{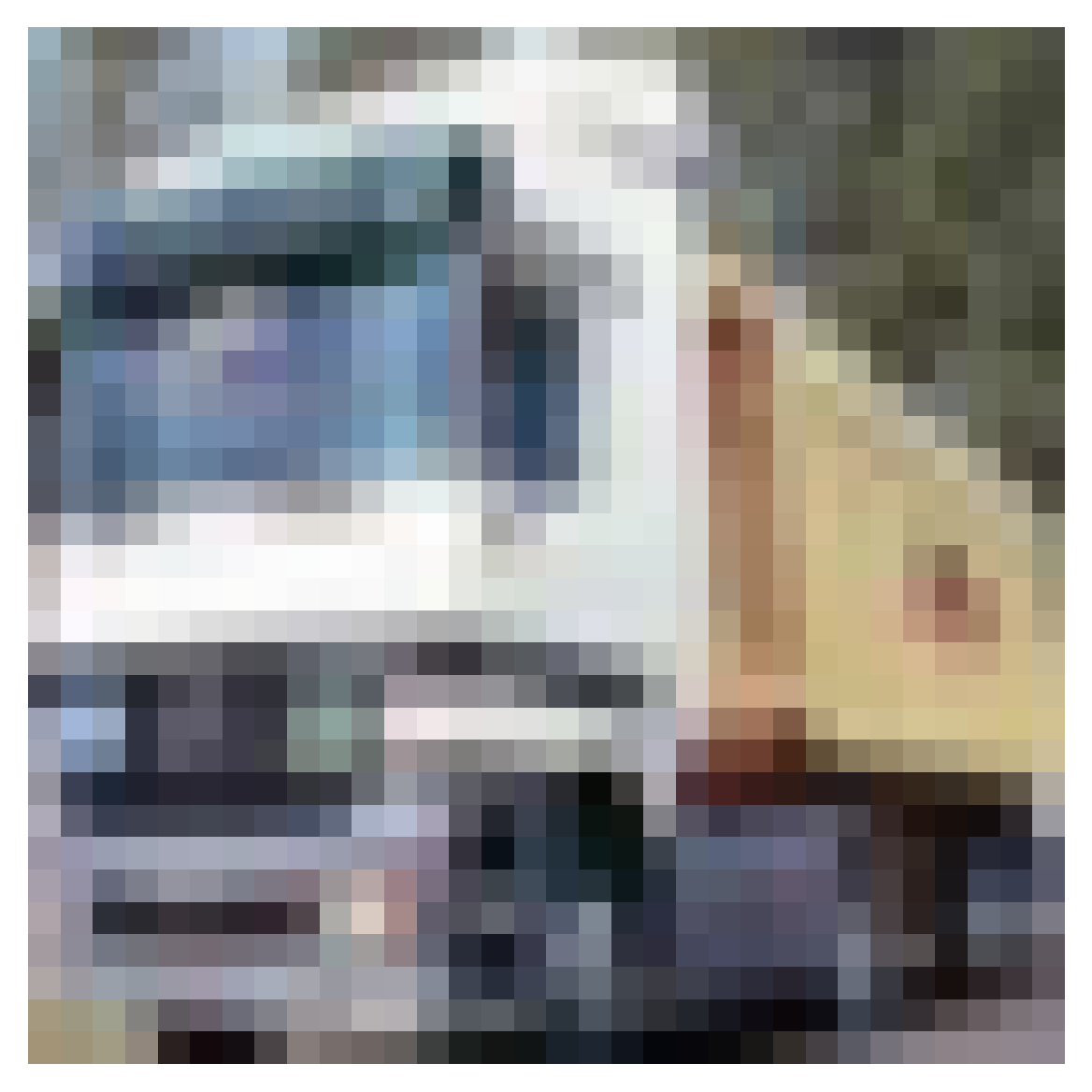} & 
    \includegraphics[width=0.16\linewidth]{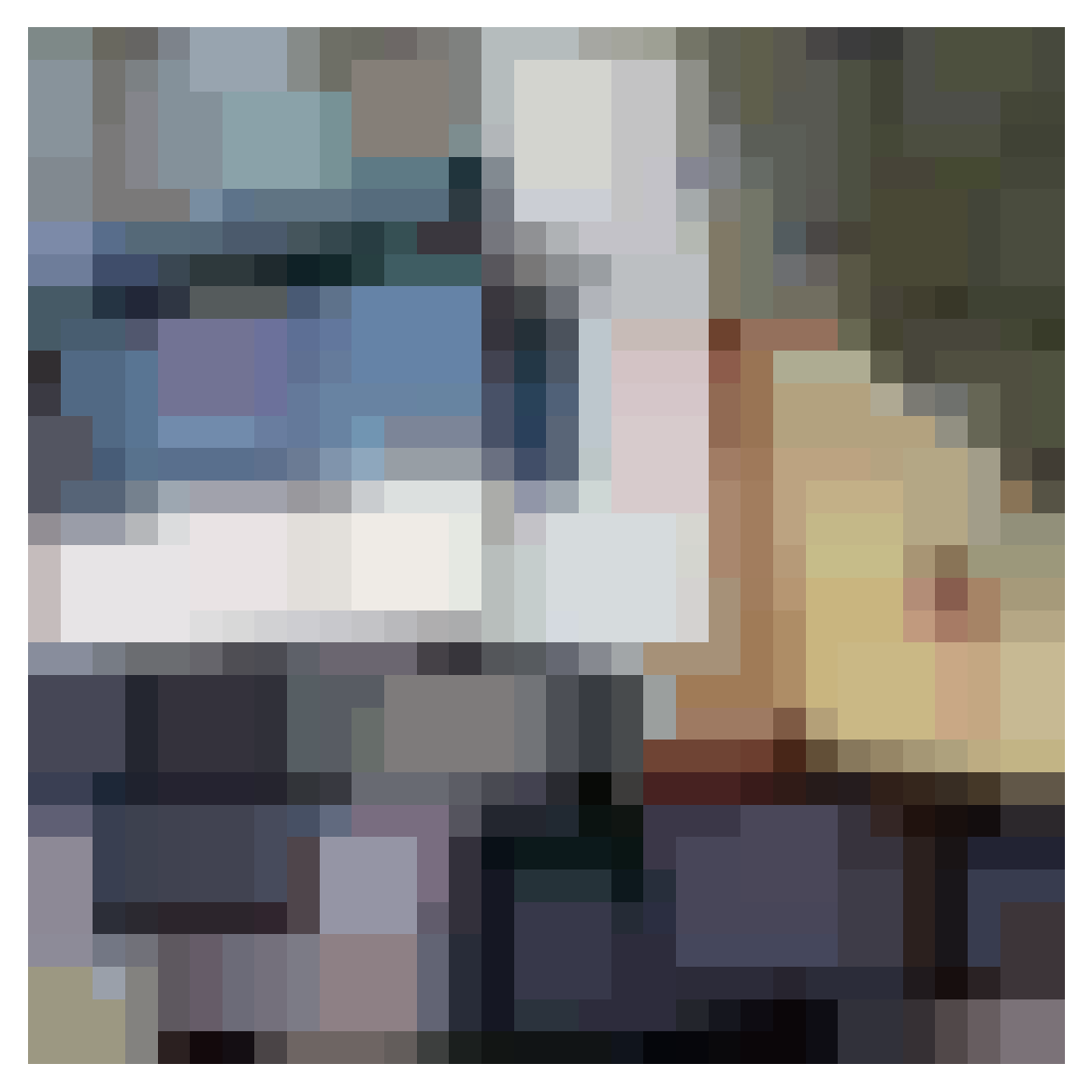} &
    \includegraphics[width=0.16\linewidth]{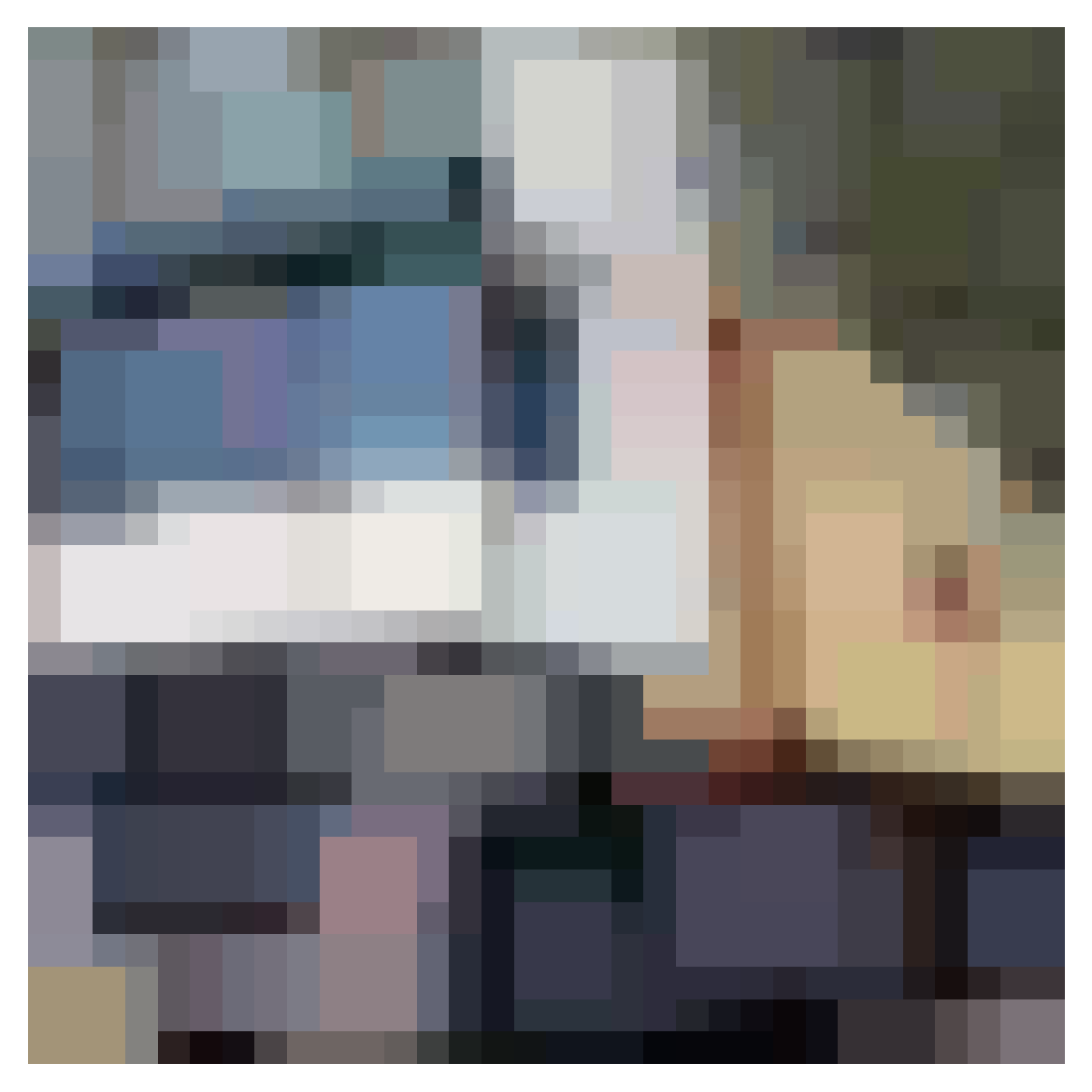} &
    \includegraphics[width=0.16\linewidth]{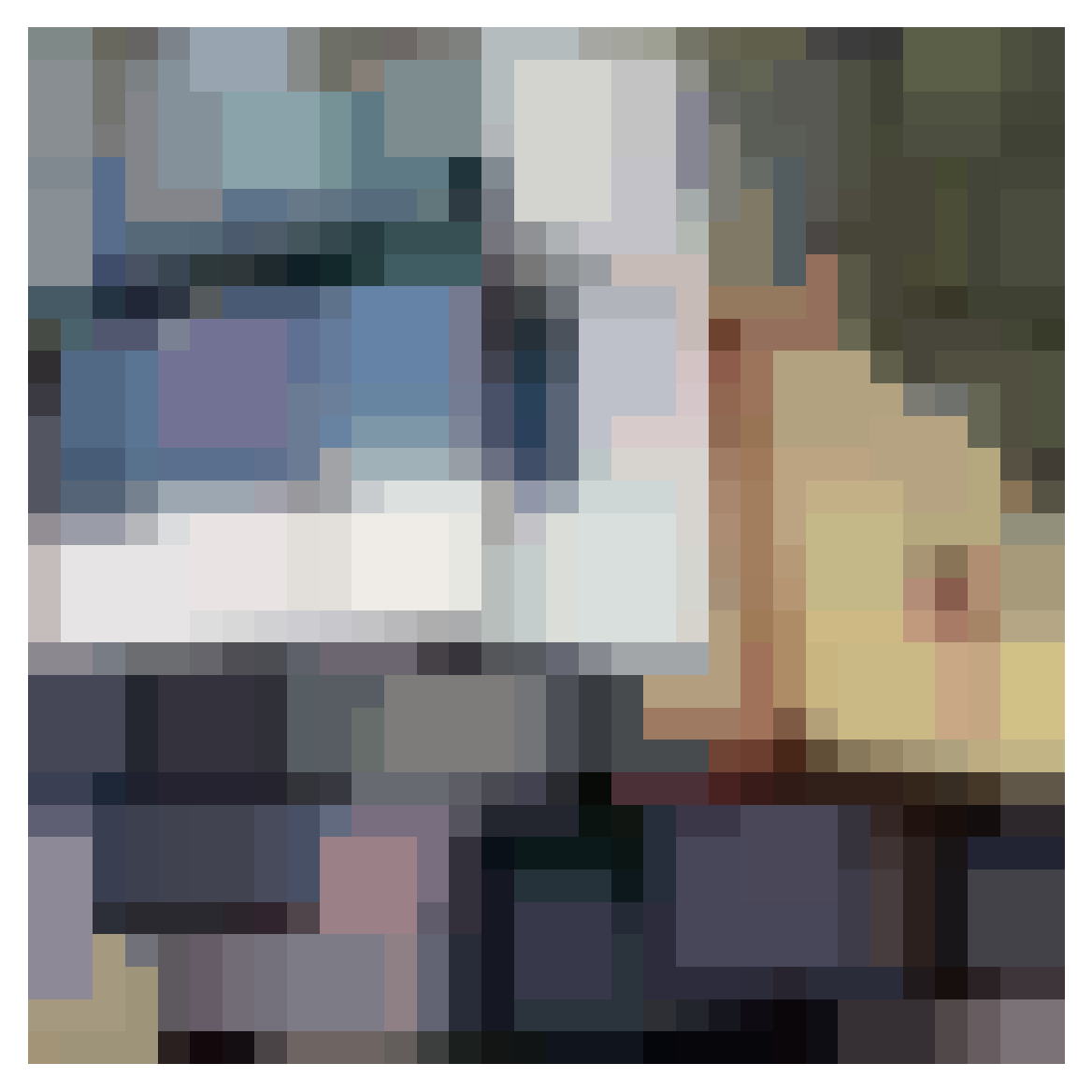} &
    \includegraphics[width=0.16\linewidth]{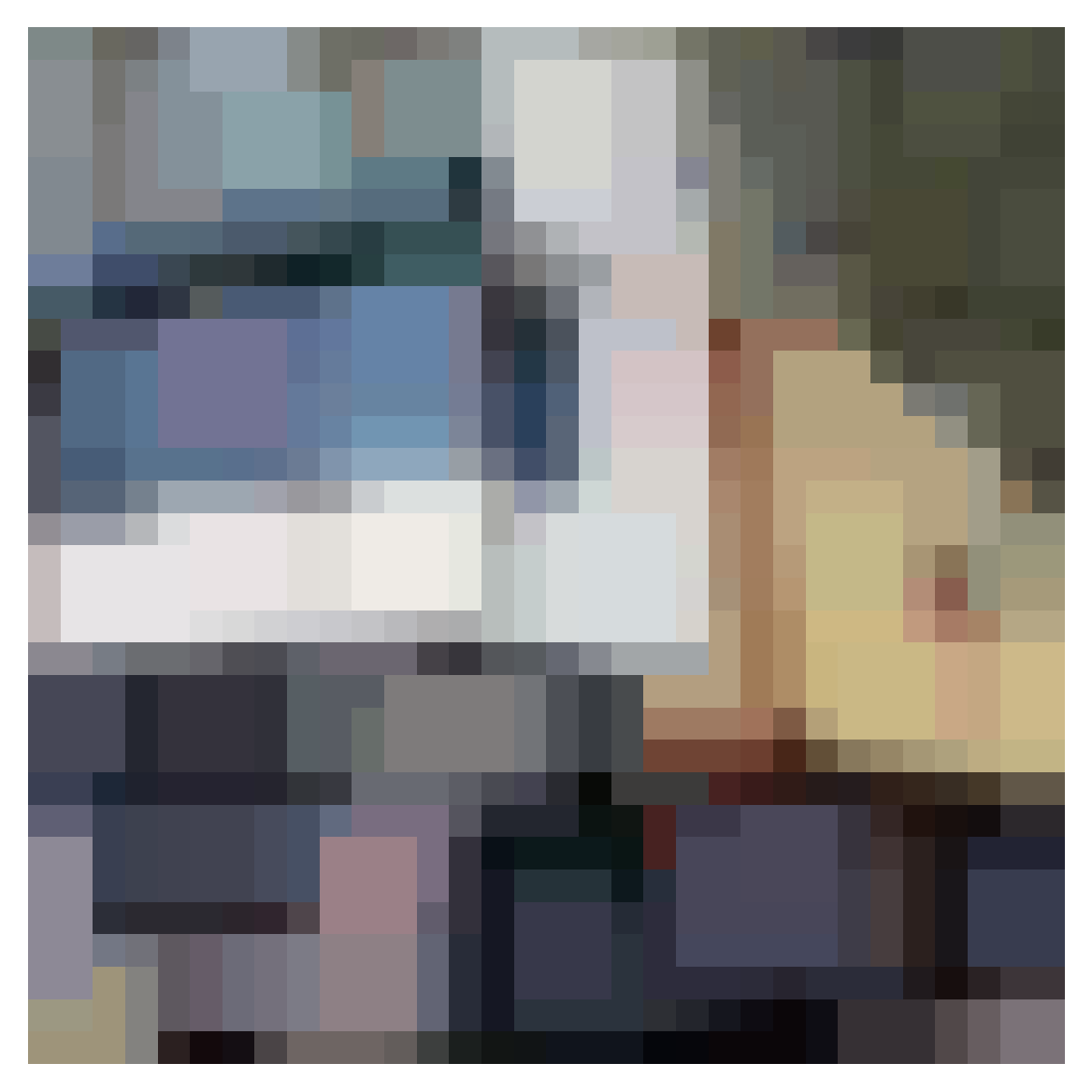} &
    \includegraphics[width=0.16\linewidth]{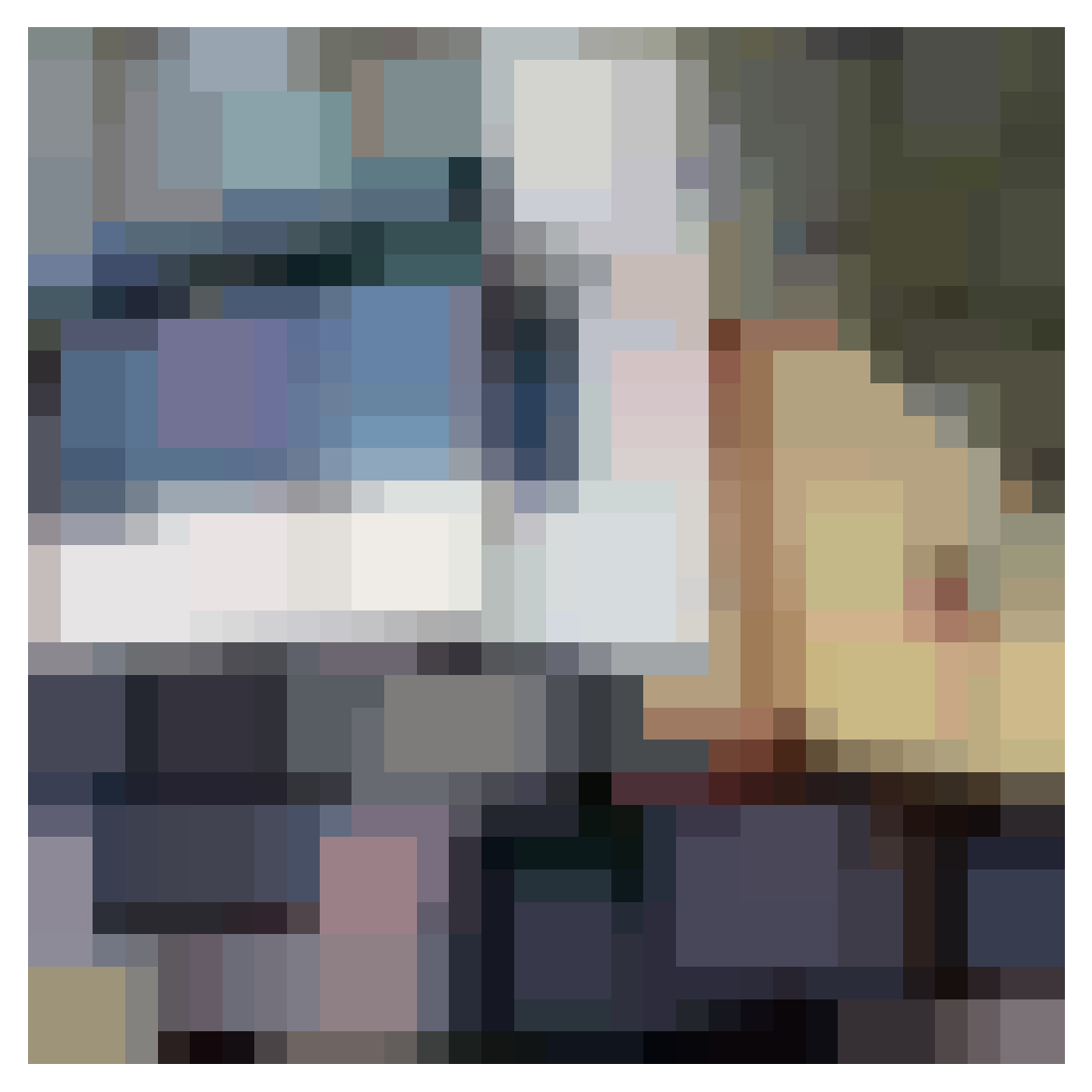} 
    \\ 
    \includegraphics[width=0.16\linewidth]{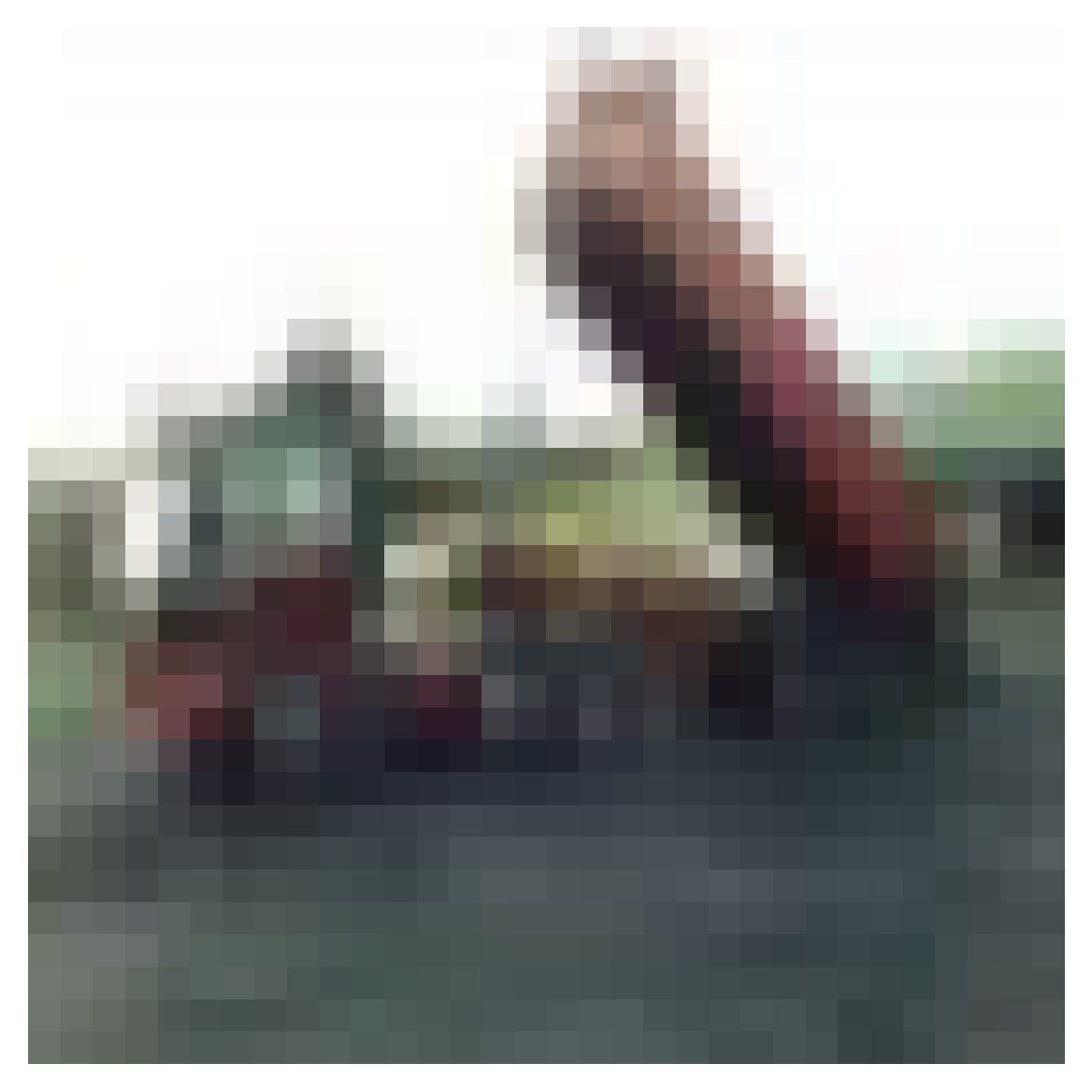} & 
    \includegraphics[width=0.16\linewidth]{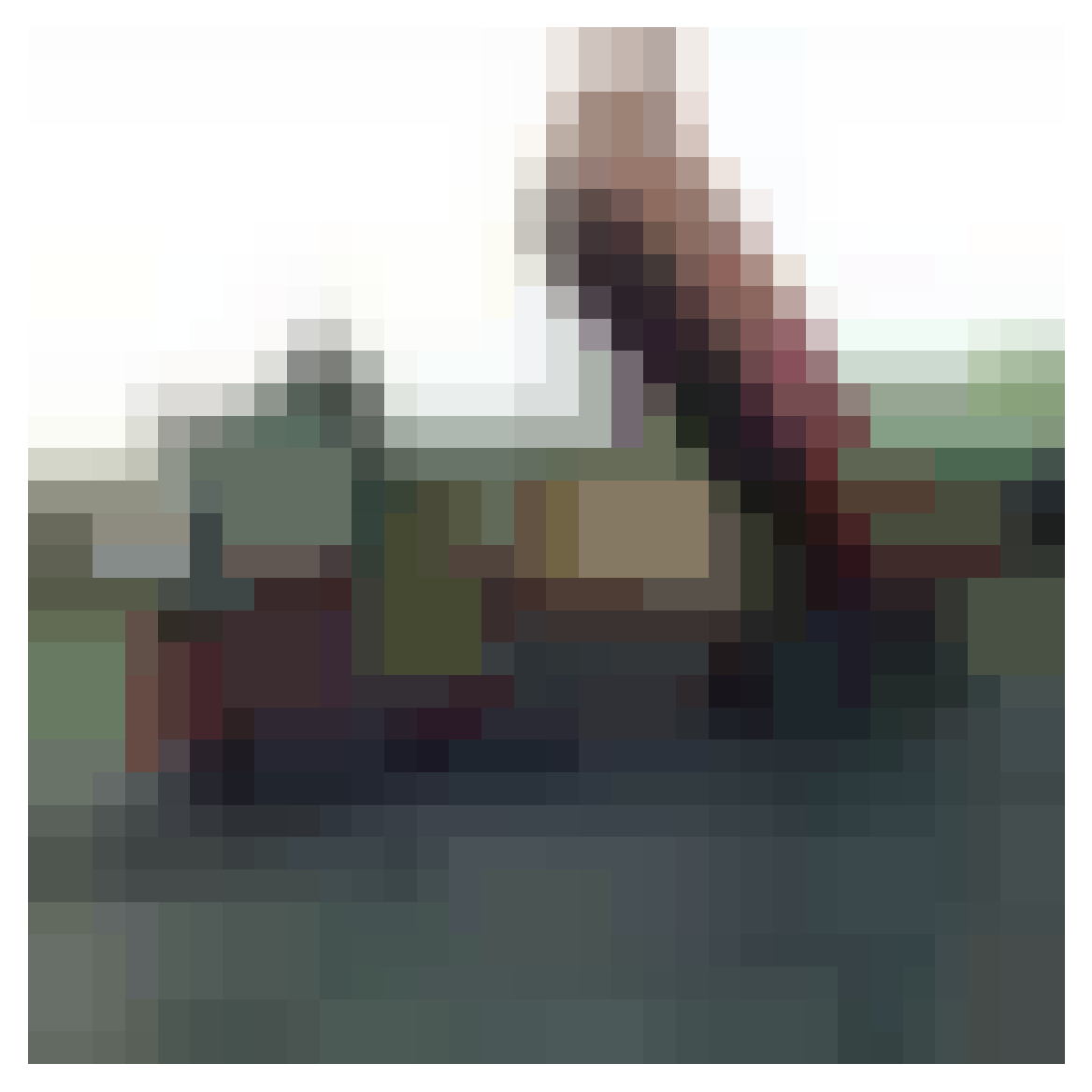} &
    \includegraphics[width=0.16\linewidth]{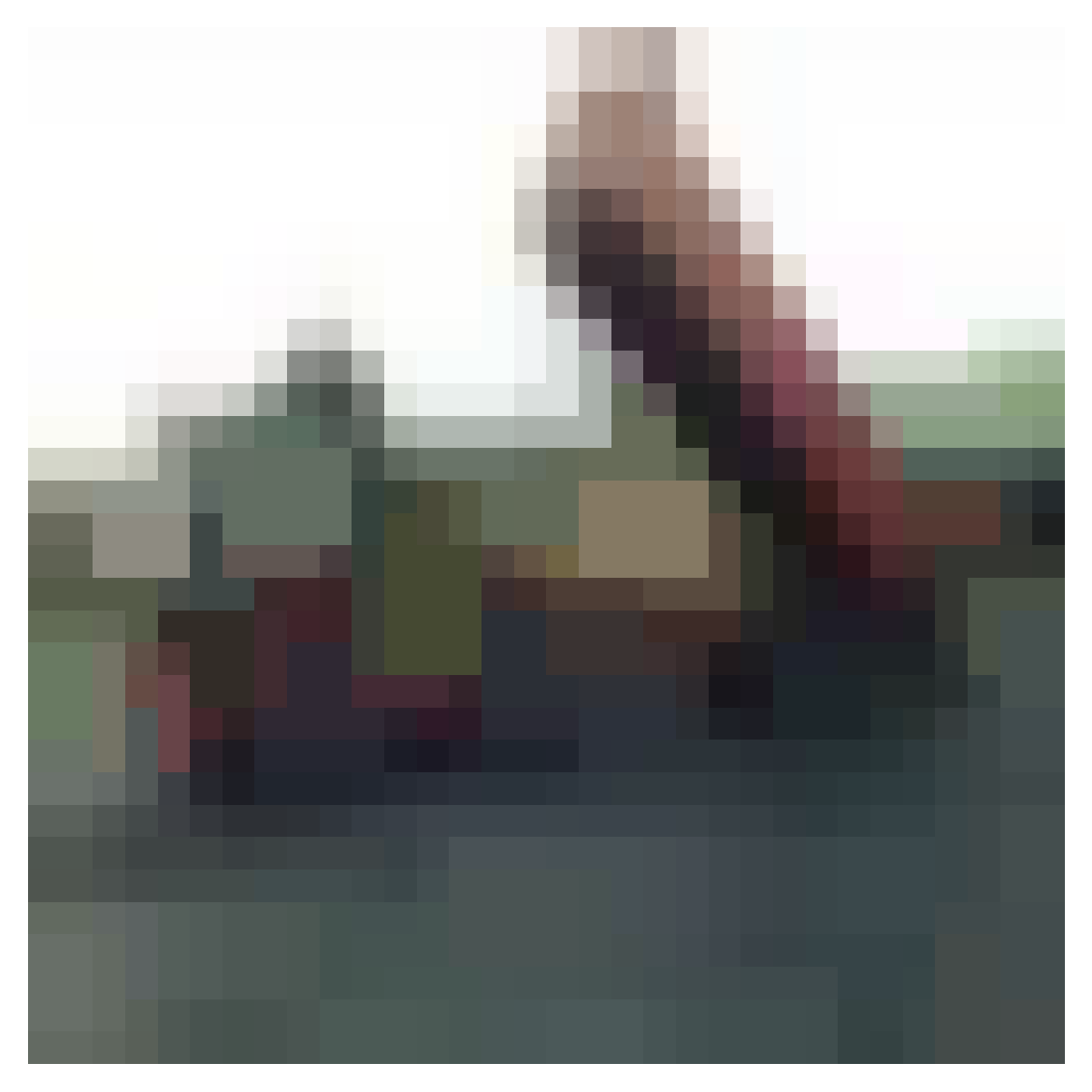} &
    \includegraphics[width=0.16\linewidth]{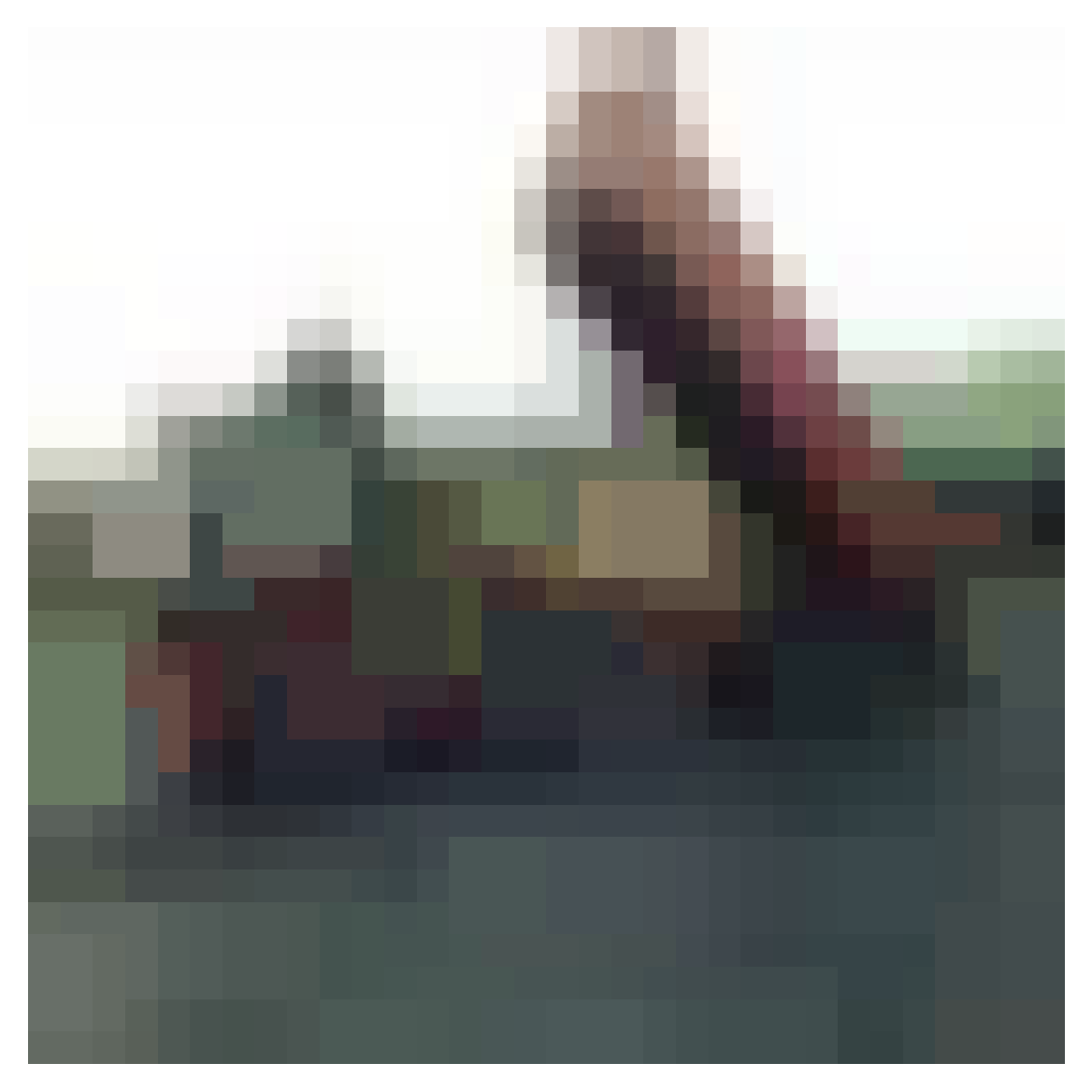} &
    \includegraphics[width=0.16\linewidth]{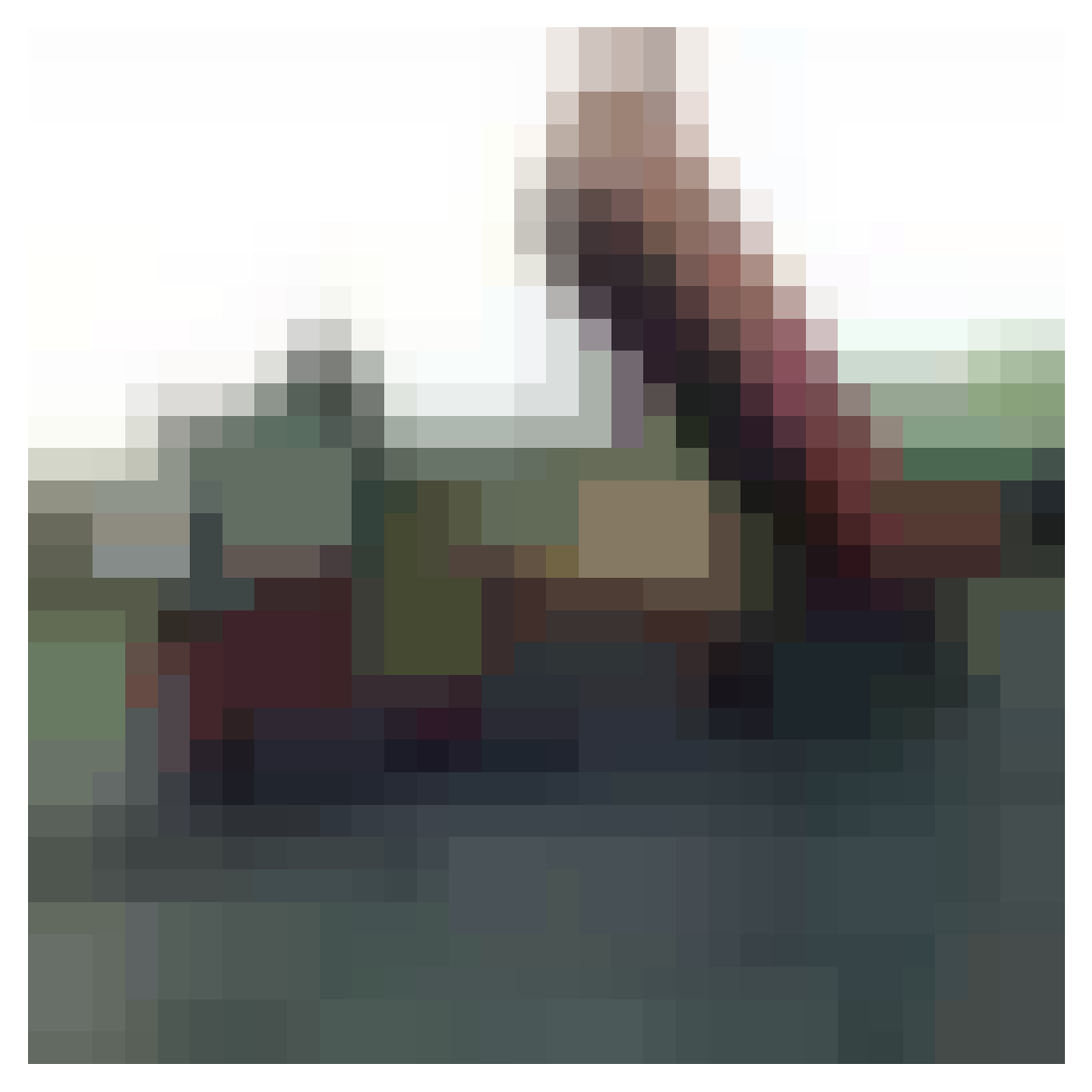} &
    \includegraphics[width=0.16\linewidth]{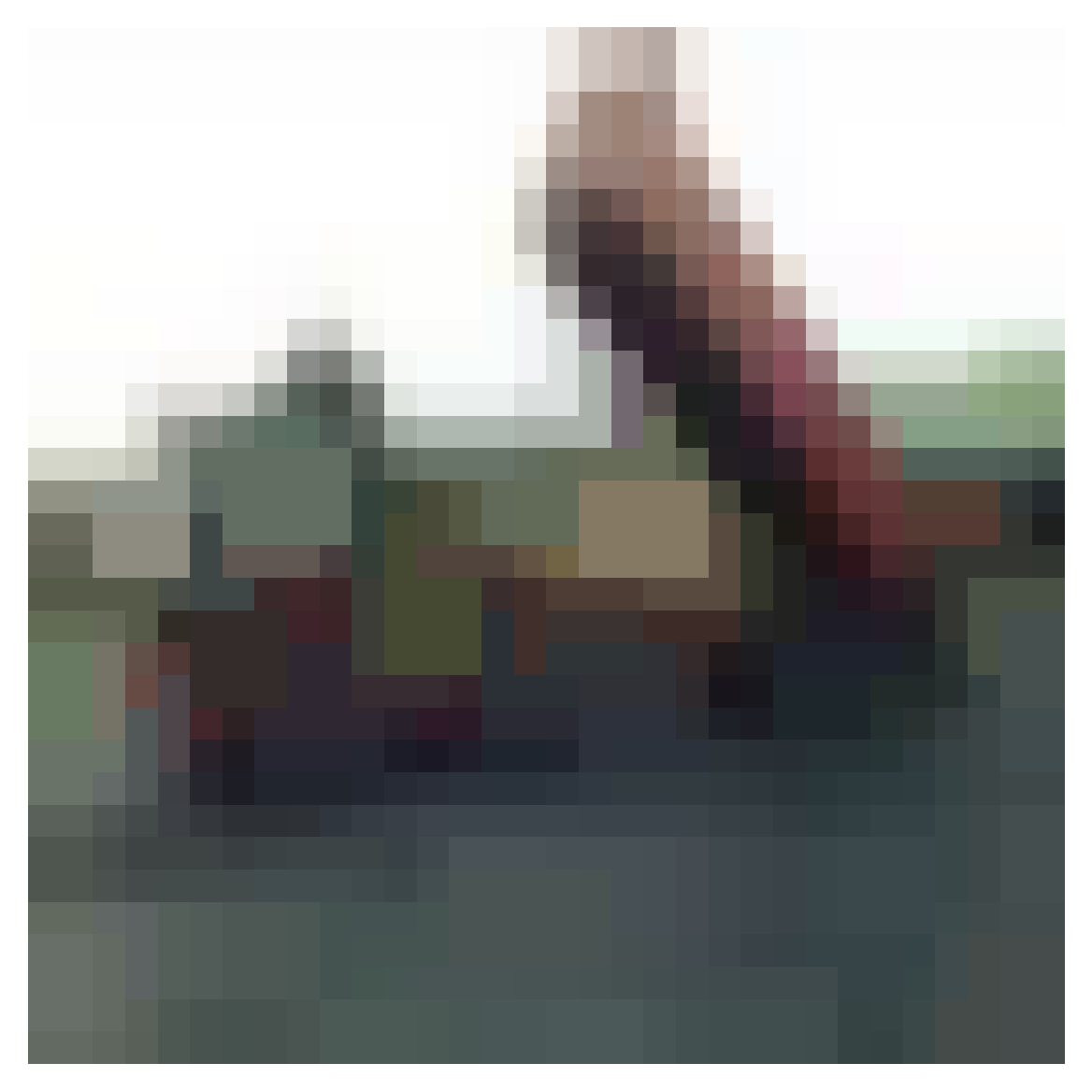} 
    \\ 
    \includegraphics[width=0.16\linewidth]{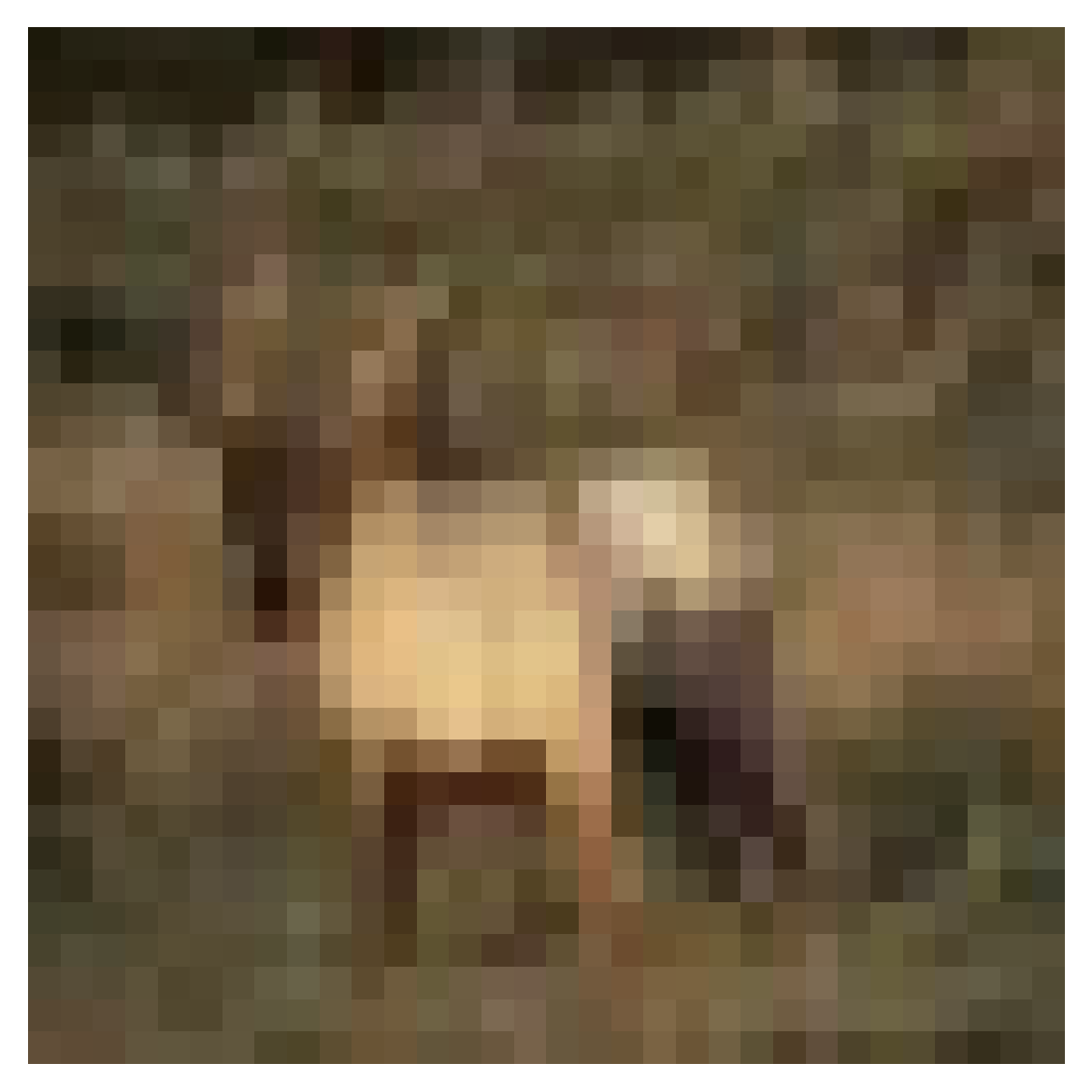} & 
    \includegraphics[width=0.16\linewidth]{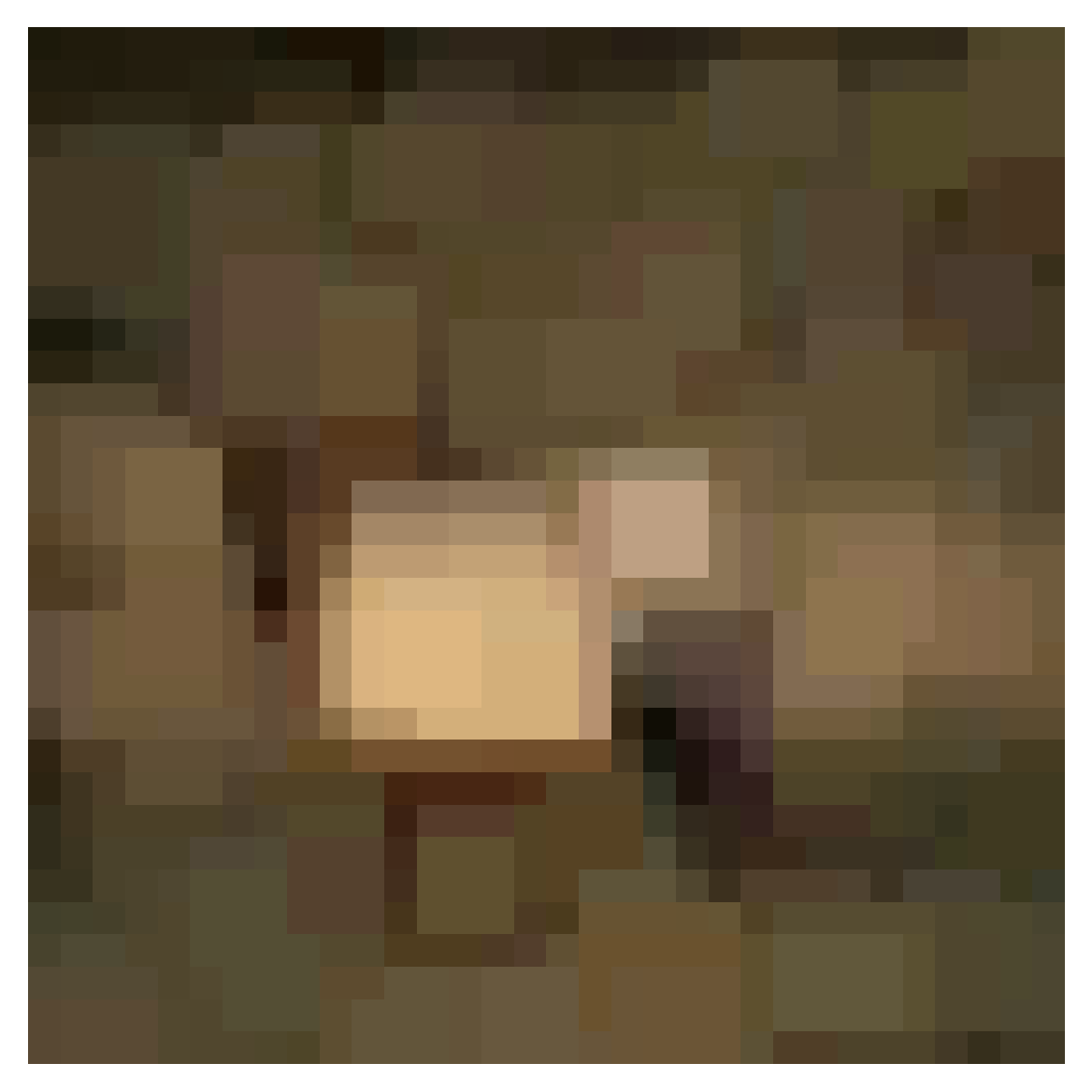} &
    \includegraphics[width=0.16\linewidth]{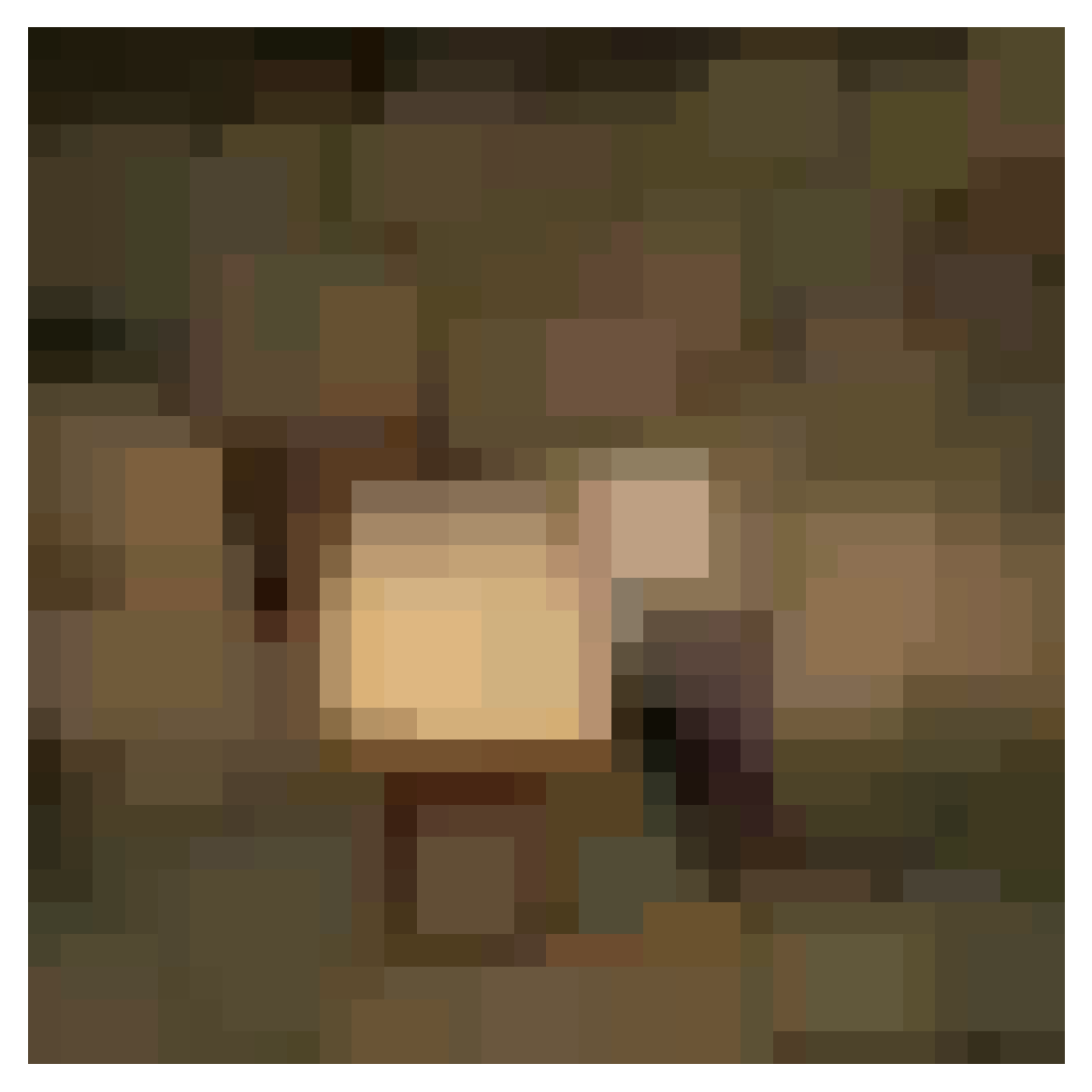} &
    \includegraphics[width=0.16\linewidth]{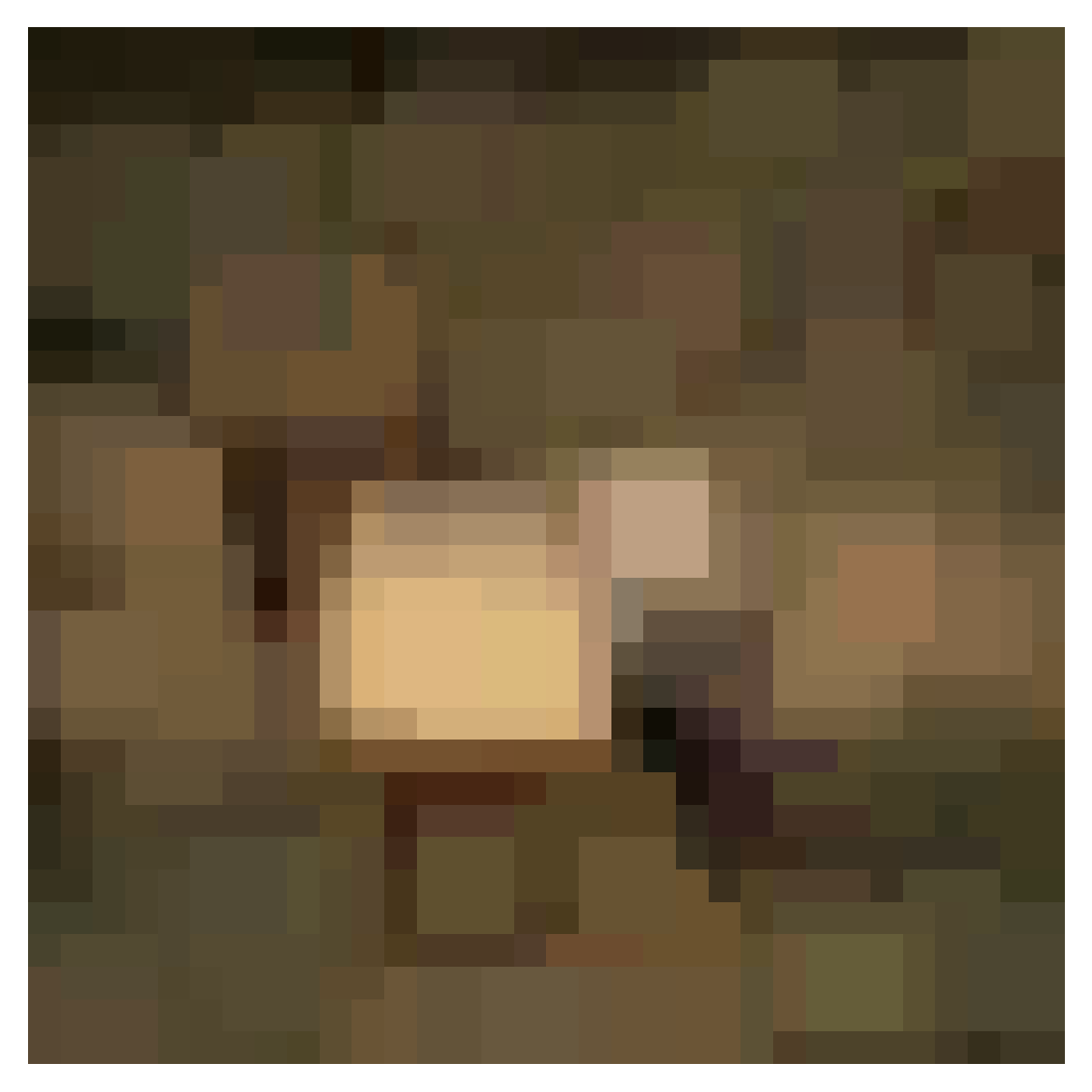} &
    \includegraphics[width=0.16\linewidth]{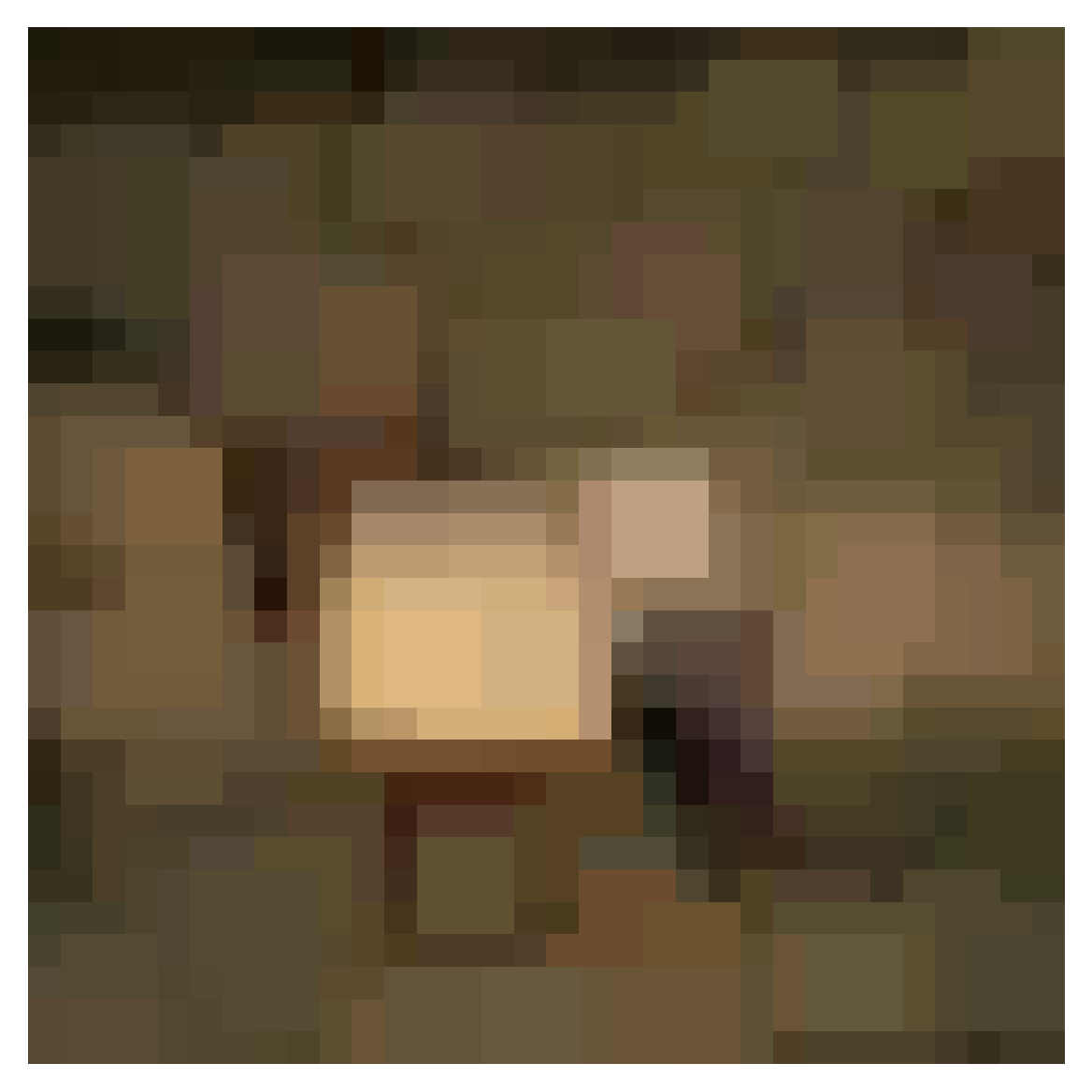} &
    \includegraphics[width=0.16\linewidth]{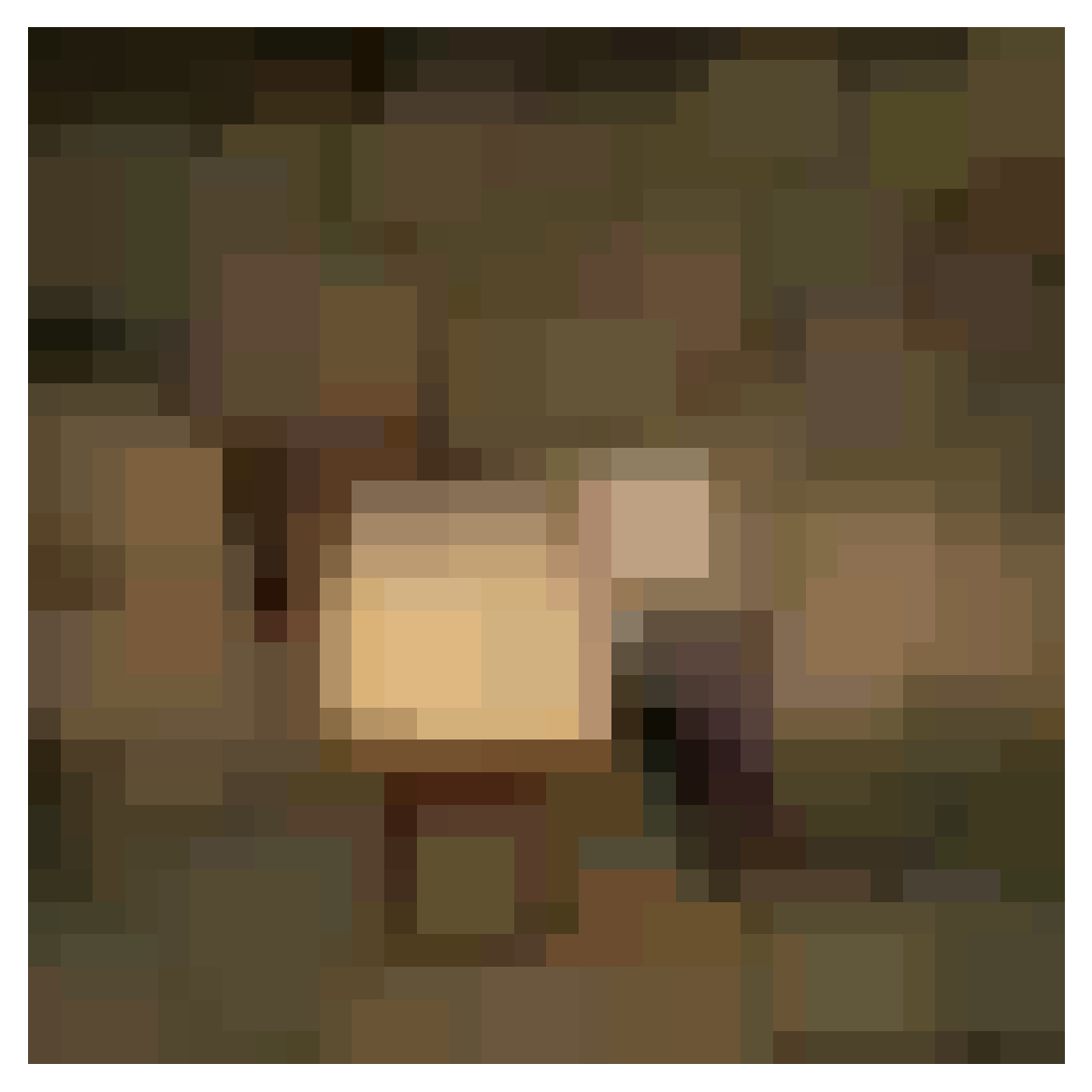} 
    \\ 
    \includegraphics[width=0.16\linewidth]{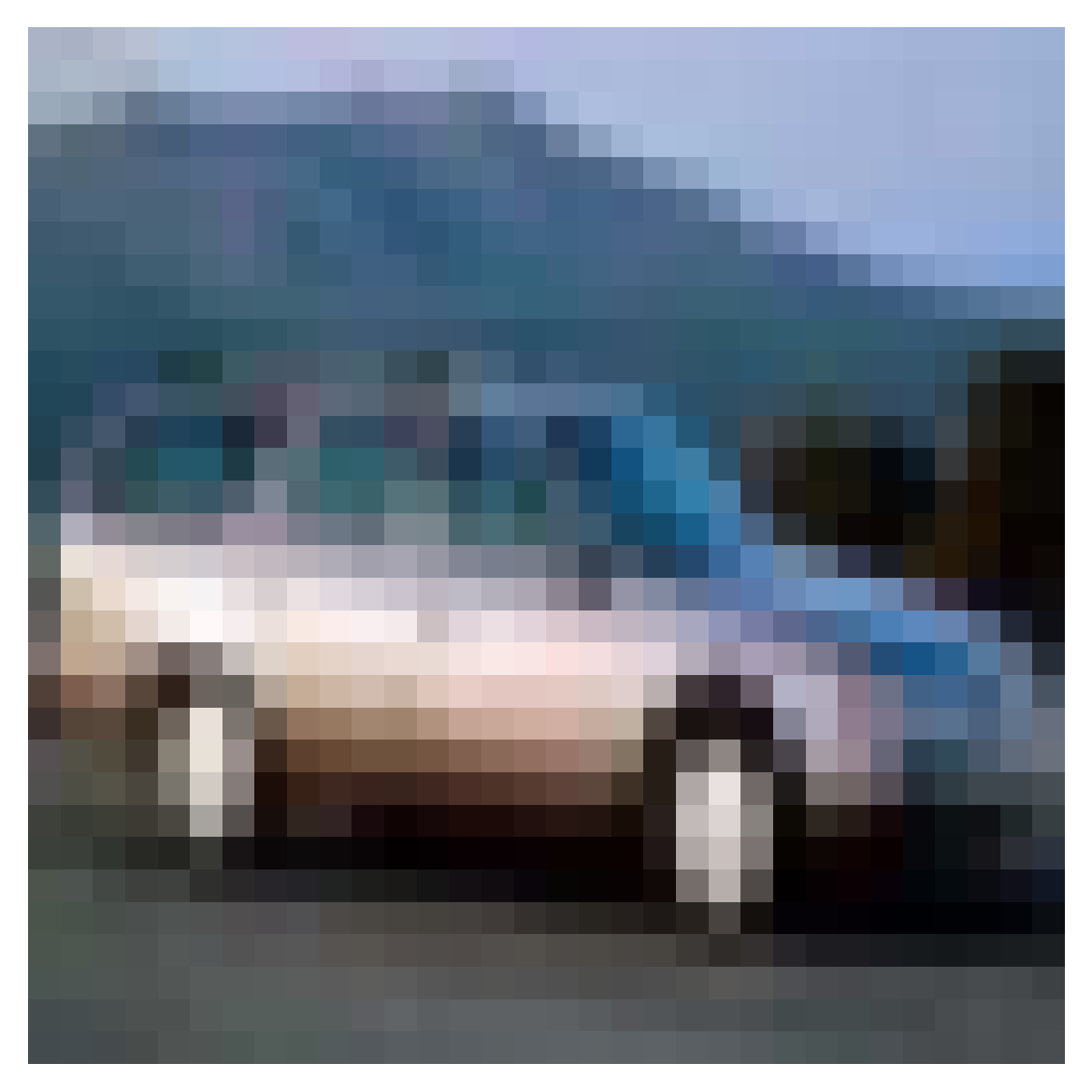} & 
    \includegraphics[width=0.16\linewidth]{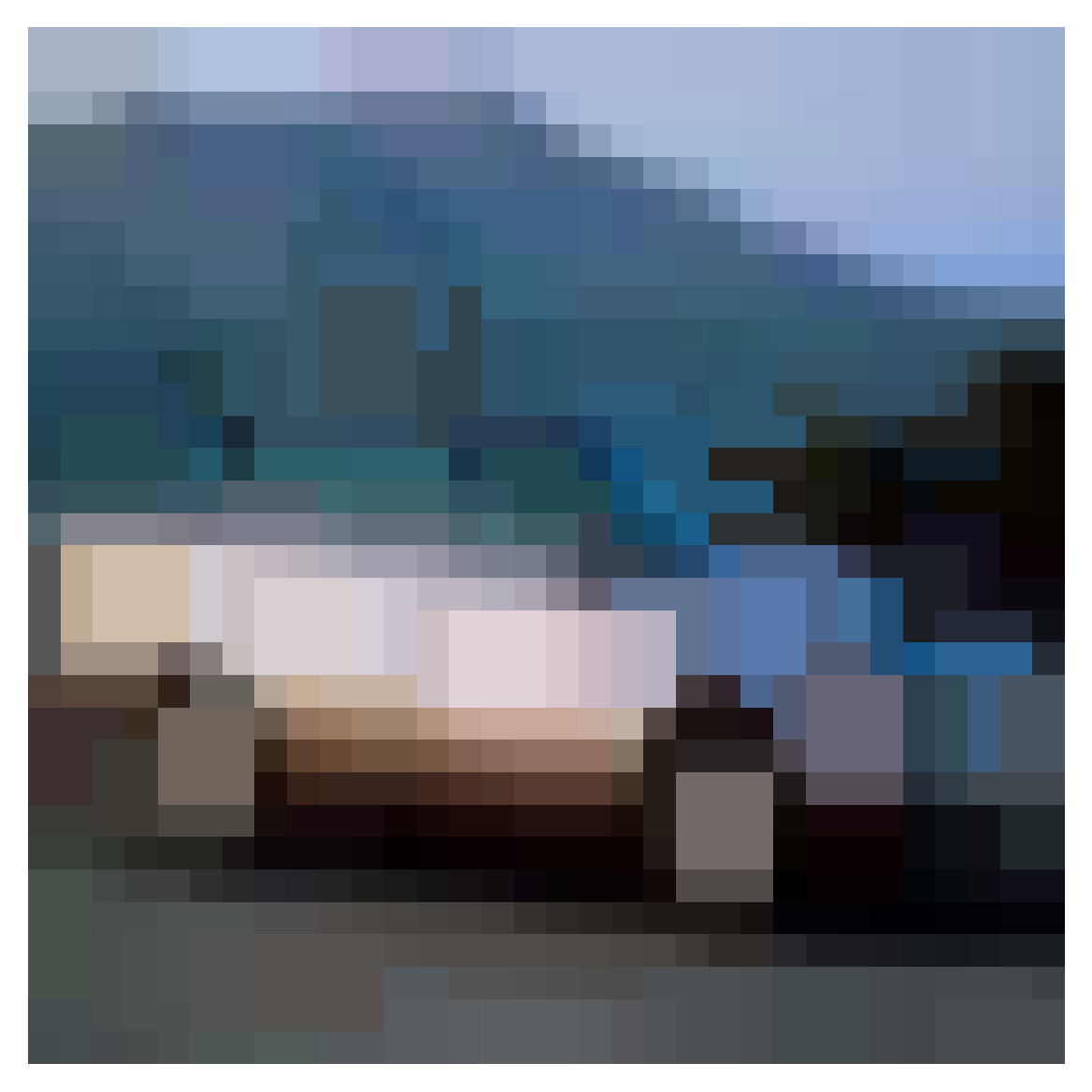} &
    \includegraphics[width=0.16\linewidth]{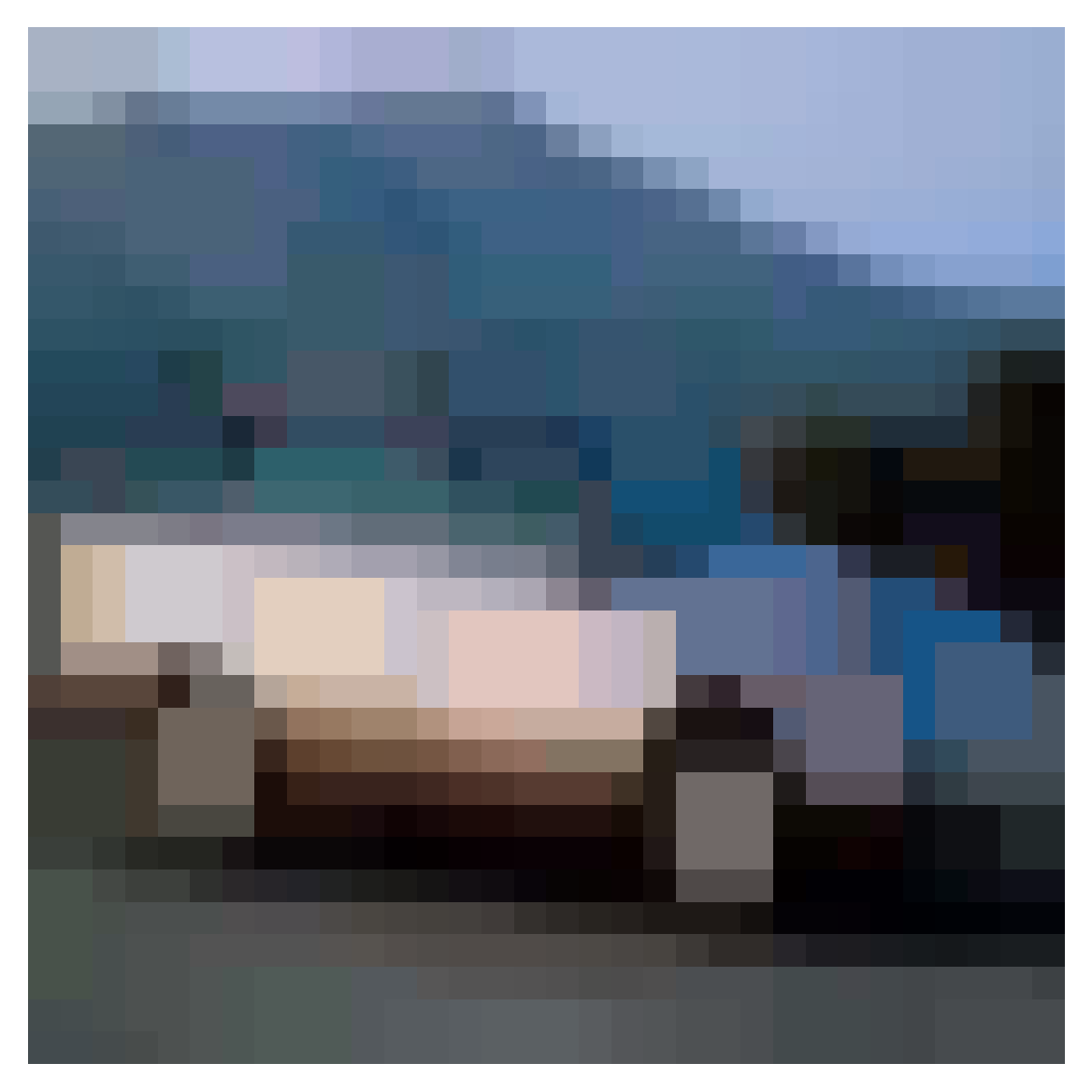} &
    \includegraphics[width=0.16\linewidth]{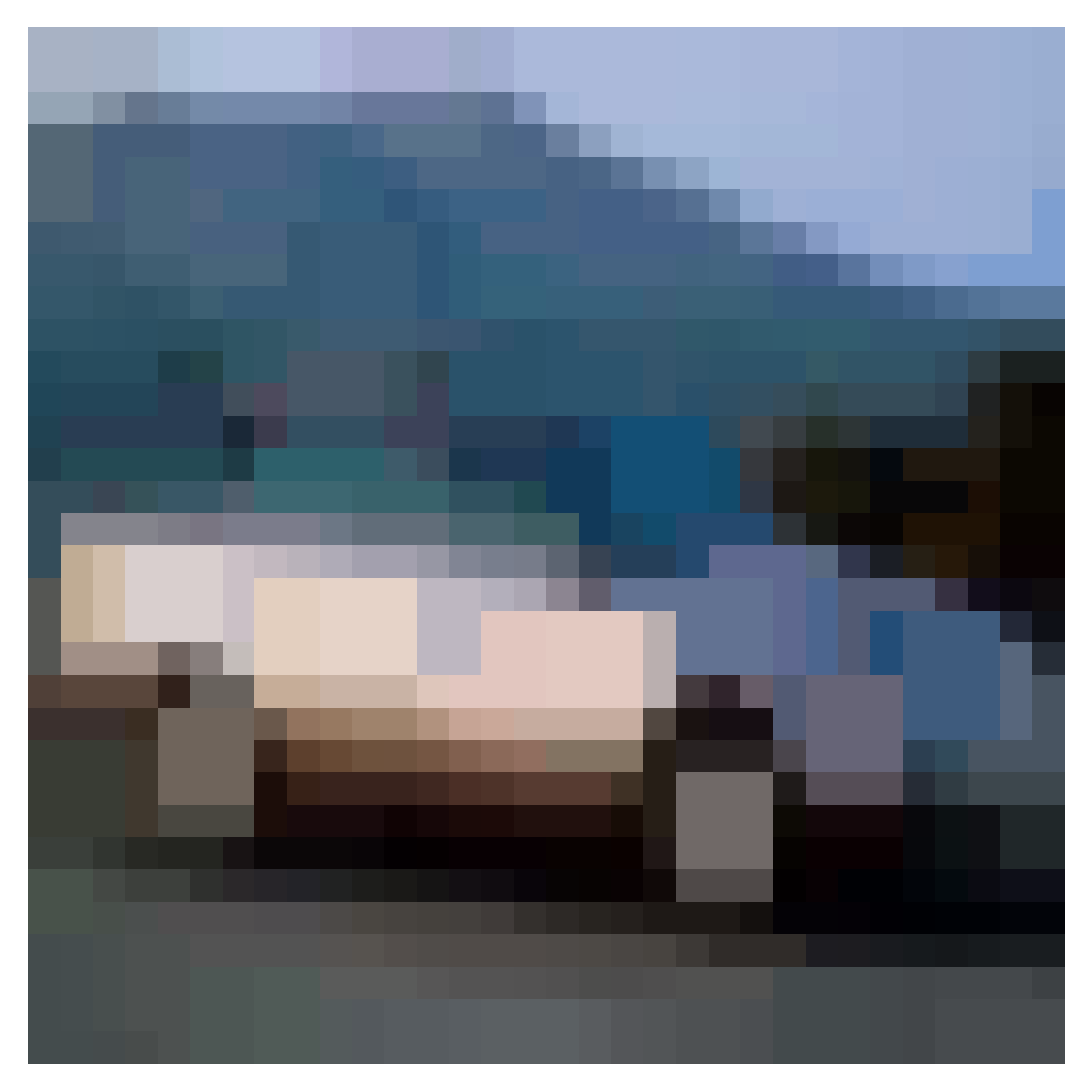} &
    \includegraphics[width=0.16\linewidth]{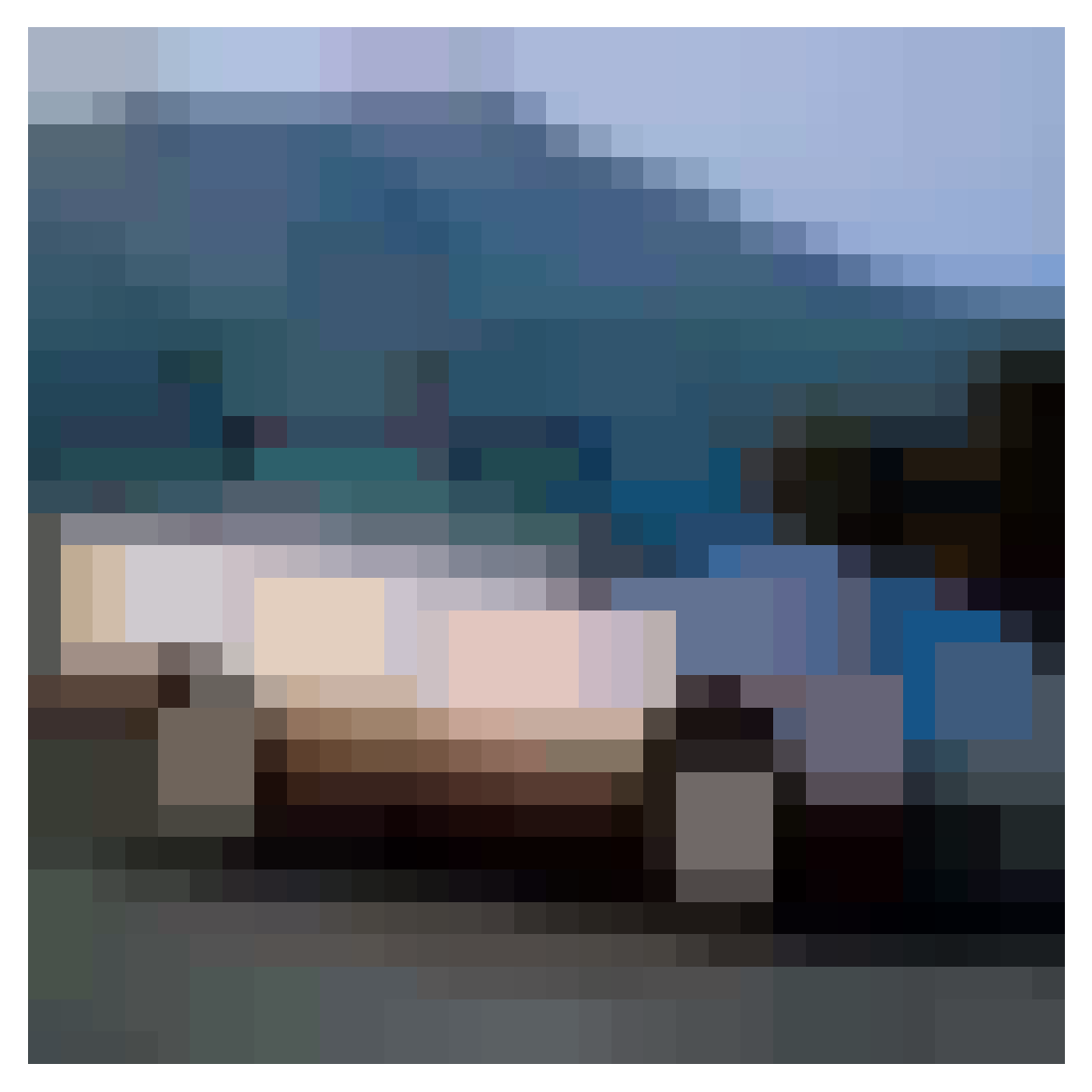} &
    \includegraphics[width=0.16\linewidth]{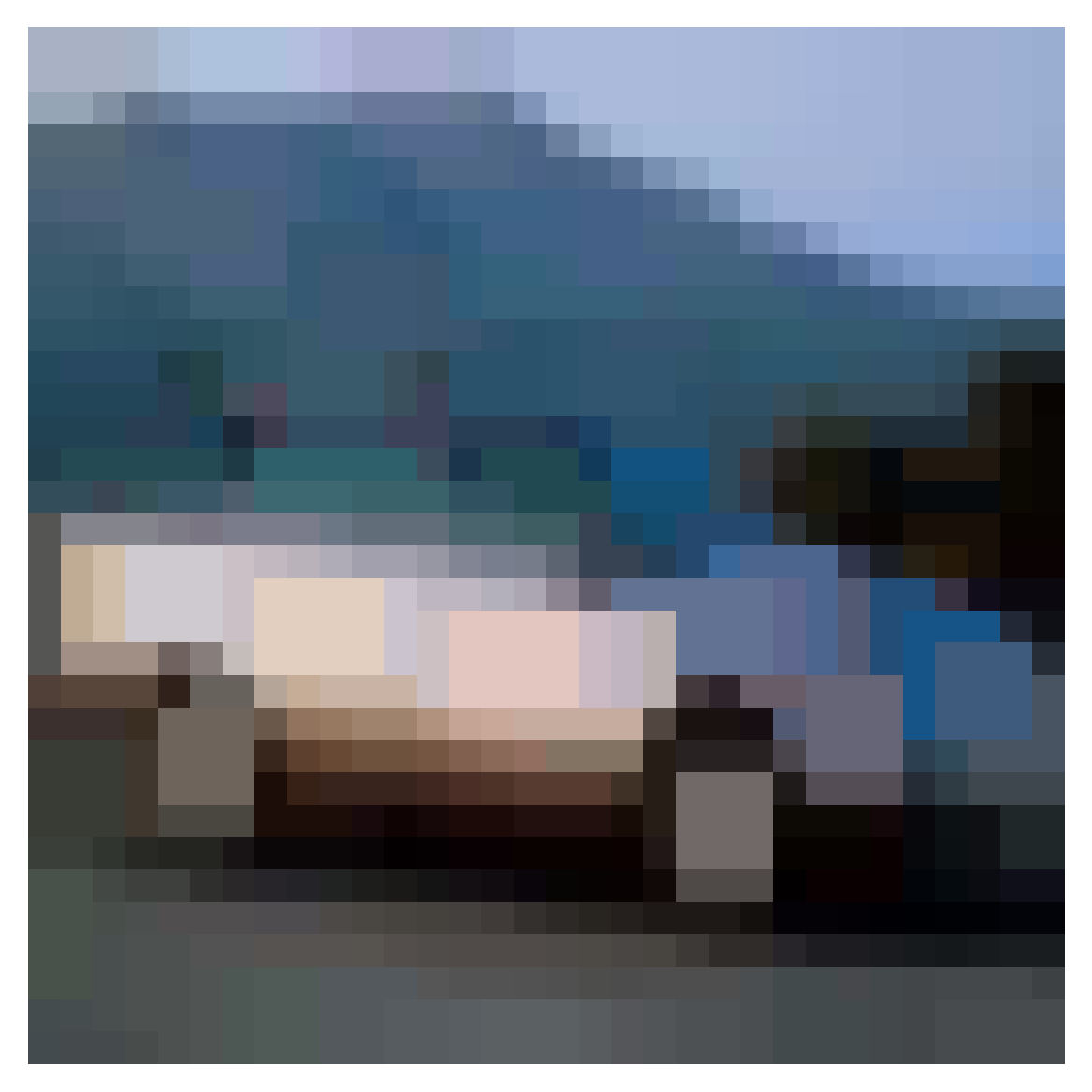} 
    \end{tabular}
    \caption{Illustrative images from the training set $\mathcal{T}$ (first column) and the corresponding opened images obtained using the reduced mappings $h_1$, $h_2$, $h_3$, the Borda rule $B$, and the Condorcet $h^*$-mapping.}
    \label{fig:cifar-images}
\end{figure}

We used the CIFAR dataset to learn the Condorcet $h$-mapping. The CIFAR dataset includes thousands of RGB color images, each of size $32 \times 32$, which are divided into training and testing sets \cite{cifar}. Although this dataset was originally designed for image classification tasks, the CIFAR images are also suitable for our purposes. Indeed, we selected the first 100 images from the CIFAR-10 training set as our training set, denoted as $\mathcal{T}_{tr} = \{\imI_1, \ldots, \imI_{100}\}$. Figure \ref{fig:cifar-images} shows some illustrative images from the training set $\mathcal{T}_{tr}$. Besides the training set, we also considered a validation set $\mathcal{T}_{val} = \{\imI_{101}, \ldots, \imI_{200}\}$ with the subsequent hundred images from the CIFAR training set. The training and validation sets $\mathbb{V}_n^{tr}$ and $\mathbb{V}_n^{val}$ consist of all $n=102,400$ colors from the hundred images in $\mathcal{T}_{tr}$ and $\mathcal{T}_{val}$, respectively.

Besides the set of color images, we considered the family of reduced mappings $\mathcal{H} = \{h_1, h_2, h_3\}$ for learning the Condorcet $h^*$-mapping \new{and the Borda rule}. The following equations give the $h$-mappings for any $\vetx = [r, g, b] \in \mathbb{V}$, where $\mathbb{V} = \{0,\frac{1}{255},\ldots,\frac{254}{255},1\}^3$ is a discrete set of RGB colors:
\begin{equation}
h_1(\vetx) = 255r+g+\frac{b}{255}, \quad
h_2(\vetx) = \frac{r}{255}+255g+b, \quad \text{and} \quad
h_3(\vetx) = r+\frac{g}{255}+255b.    
\end{equation}
We would like to remark that the $h$-mappings $h_1$, $h_2$, and $h_3$ yield the lexicographic orderings $R-G-B$, $G-B-R$, and $B-R-G$, respectively. These three $h$-mappings have been chosen for illustrative purposes in this paper. In practical scenarios, the $h$-mappings can be defined using supervised or unsupervised approaches, as noted in \cite{Lezoray2021MathematicalOrderings,velasco-forero14}.

\begin{figure}[t]
    \centering
    \includegraphics[width=0.6\linewidth]{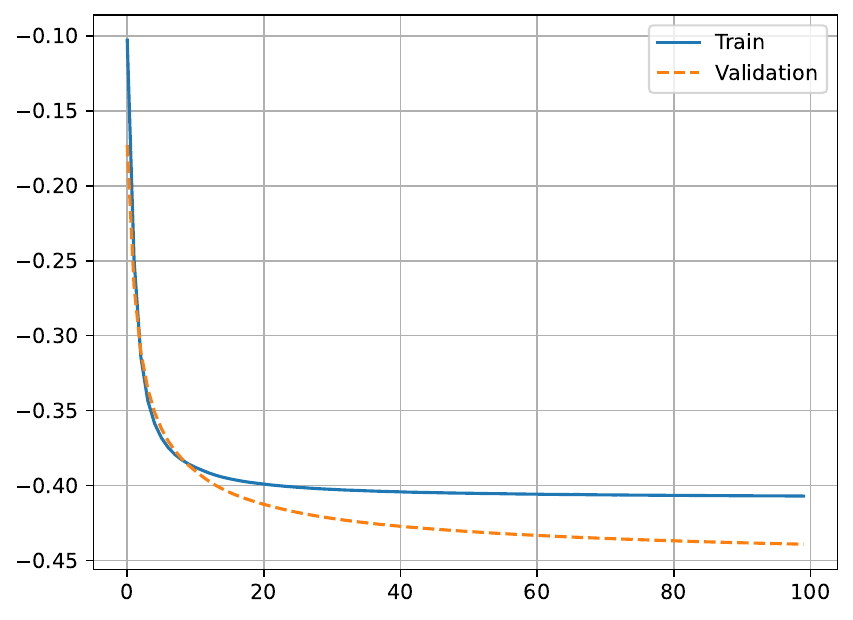}     \caption{The loss over epochs during the training of the Condorcet $ h^*$-mapping.}
    \label{fig:loss}
\end{figure}

We adopted a multilayer perceptron (MLP) network with an architecture 3-64-1 for our computational experiments. In other words, the MLP network consists of a single hidden layer containing 64 neurons that utilize the ReLU activation function, along with a single output neuron that has no bias and employs the identity function as its activation function (effectively meaning no activation function is used for the output neuron). The MLP network inputs an RGB color value \(\vetx = [r, g, b]\) and produces a score value as the output. The optimization problem outlined in \eqref{eq:SoftCondorcetOptimization}, with the loss function defined by \eqref{eq:soft-loss} with $\tau=1$, was solved using the Adam optimizer with the default parameters from Keras, version 3.6.0. The training was conducted over a fixed number of 100 epochs, using batches of size 1024. We used the validation set $\mathcal{V}_n^{val}$ to assess the generalization capability of the proposed MLP network. Figure \ref{fig:loss}a) plots the loss values against the number of epochs. 
%%%%%%%%%%%%%%%%%%%%%%%%%%%%%%%%%%%%%%%%%%%%%%%%%%%%%%%%%%%%%%
%%%%%%%%%%%%%%%%%%%%%%%%%%%%%%%%%%%%%%%%%%%%%%%%%%%%%%%%%%%%%%%%%%%
Note that the loss levels off after 60 epochs, indicating that the solver has probably reached a minimum of the loss function. Also, the loss curves for the training and validation sets are similar, confirming the proposed network's ability to generalize well. The trained MLP network corresponds to the Condorcet $h^*$-mapping. 

Figure \ref{fig:cifar-images} illustrates the opening operation, denoted as $\gamma_S^h$, performed with a $3 \times 3$ square structuring element, using the reduced ordering approaches, \new{including the proposed approach, and the Borda rule}. Specifically, the first five color images from the training set $\mathcal{T}_{tr}$ were processed with the morphological operation defined by the mappings $h_1$, $h_2$, and $h_3$. The resulting images are displayed in the second, third, and fourth columns of Figure \ref{fig:cifar-images}. \new{The last two columns showcase the opening obtained using the Borda rule $B$ and the Condorcet $h^*$-mapping, respectively}. The images produced by \new{Borda rule and} the Condorcet $h^*$-mapping can be interpreted as a consensus generated from the other three morphological approaches. 

\begin{figure}[t]
    \centering
    \begin{tabular}{cc}
    a) Boxplot of the irregularity index & b) Diagram of a hypothesis test. \\
    \includegraphics[width=0.6\linewidth]{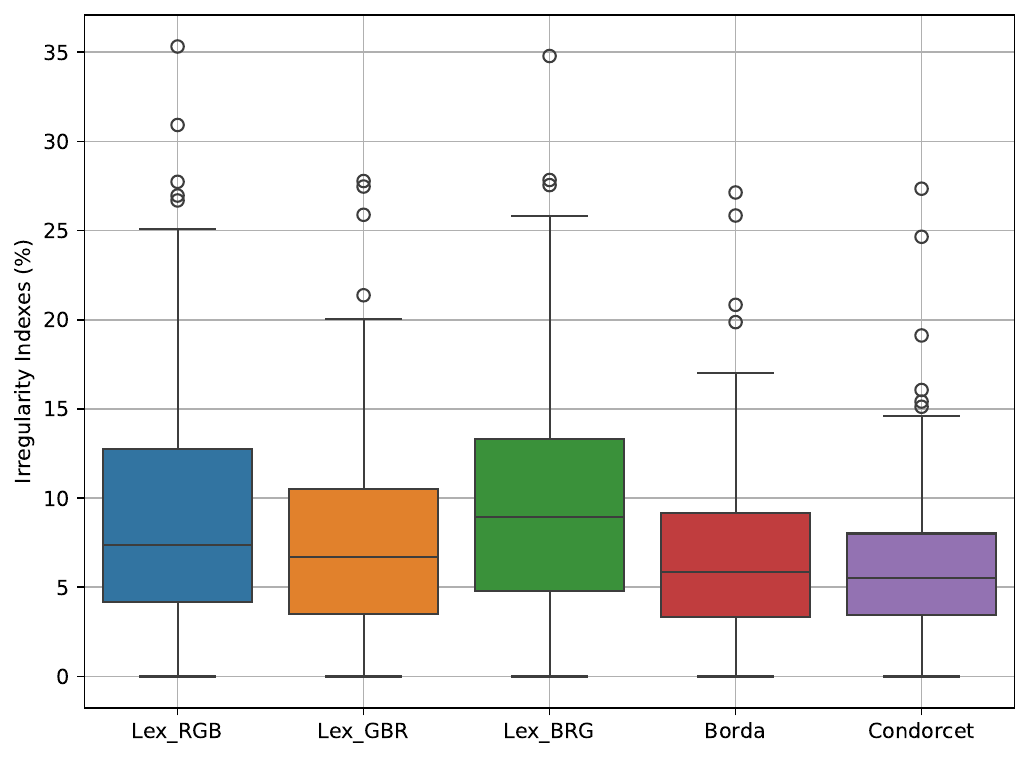} &
    \includegraphics[width=0.4\linewidth]{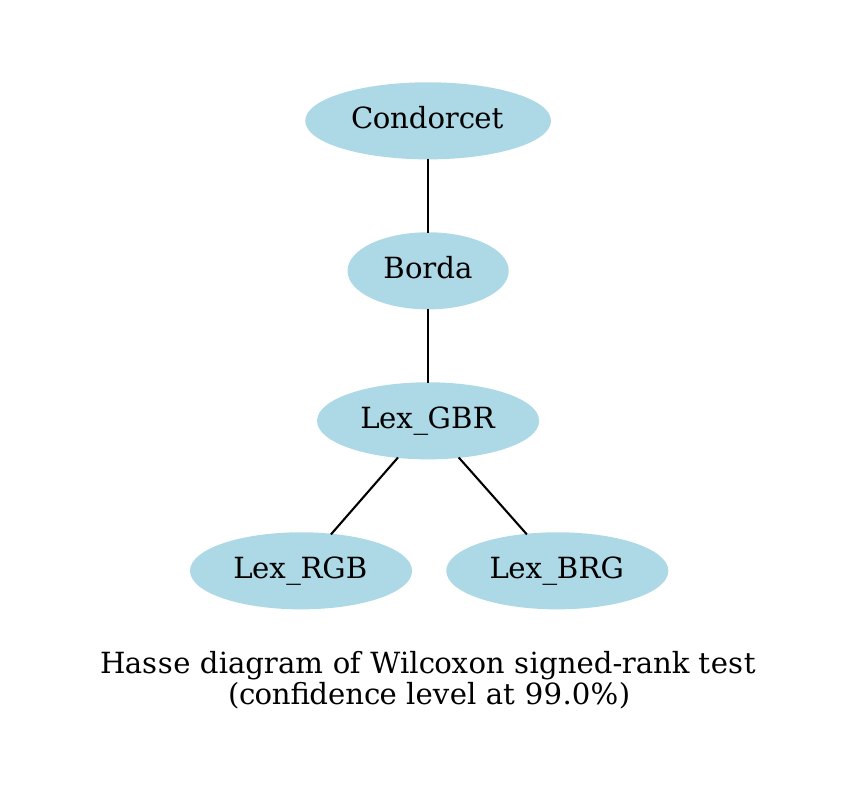}
    \end{tabular}
    \caption{Boxplot of the global irregularity values and the Hasse diagram of pair-wise Wilcoxon signed-rank test with confidence level at $99\%$.}
    \label{fig:irregularity}
\end{figure}

We would like to remark that the morphological operation does not introduce ``\textit{false colors}'' because we use a reduced ordering combined with a look-up table. Moreover, we observed that the images obtained using the Condorcet $h^*$-mapping are generally more regular than those obtained using other approaches. \new{Accordingly, Figure \ref{fig:irregularity}a) presents a boxplot illustrating the irregularity index values, $\Phi_1^g\big(\imI_k,\gamma_S^{\star}(\imI_k)\big)$, calculated for all images $\imI_k \in \mathcal{T}_{val}$. Here, $\Phi_1^g$ denotes the global irregularity index, as detailed in \cite{Valle2022IrregularityOperators}. Notably, the irregularity index values generated by the Condorcet approach display the least variability and are generally lower than those produced by the lexicographic methods and the Borda rule. We confirmed this observation through a Wilcoxon signed-rank test, conducted with a confidence level of 99\%. The Hasse diagram shown in Figure \ref{fig:irregularity}b) summarizes the results of the pair-wise Wilcoxon hypothesis test comparing the irregularity indexes of approaches \cite{weise15}. In the Hasse diagram, the models positioned at the top statistically produced the lowest irregularity index values.}

%%%%%%%%%%%%%%%%%%%%%%%%%%%%%%%%%%%%%%%%%%%%%%%%%%%%%%%%%%%%%%
%%%%%%%%%%%%%%%%%%%%%%%%%%%%%%%%%%%%%%%%%%%%%%%%%%%%%%%%%%%%%%%%%%%

\begin{figure}[t]
    \centering
    \begin{tabular}{ccc}
    $\imJ$ & $\phi_S^{B}(\imJ)$ & $\phi_S^{h^*}(\imJ)$ \\
    \includegraphics[width=0.32\linewidth]{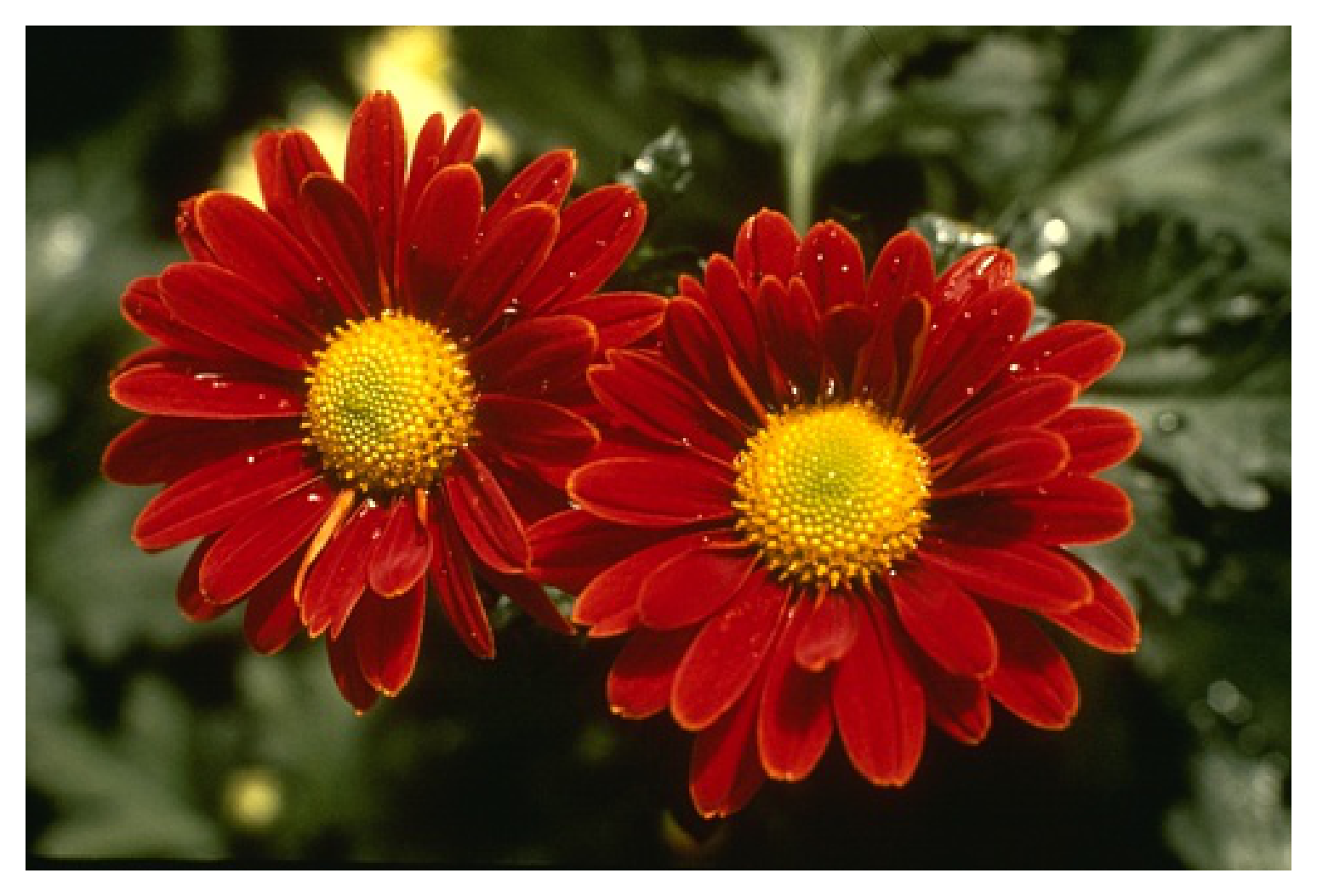} & 
    \includegraphics[width=0.32\linewidth]{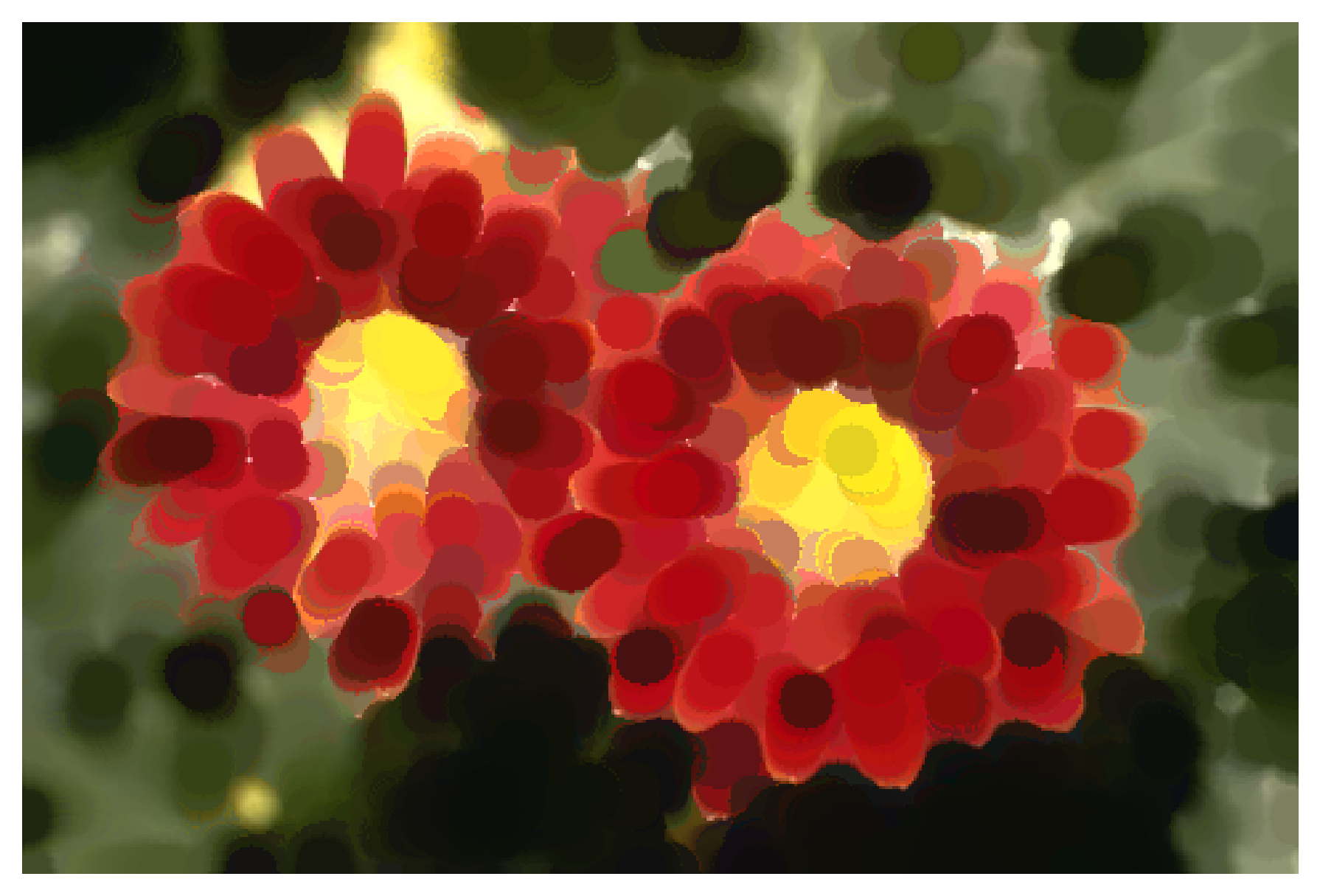} & 
    \includegraphics[width=0.32\linewidth]{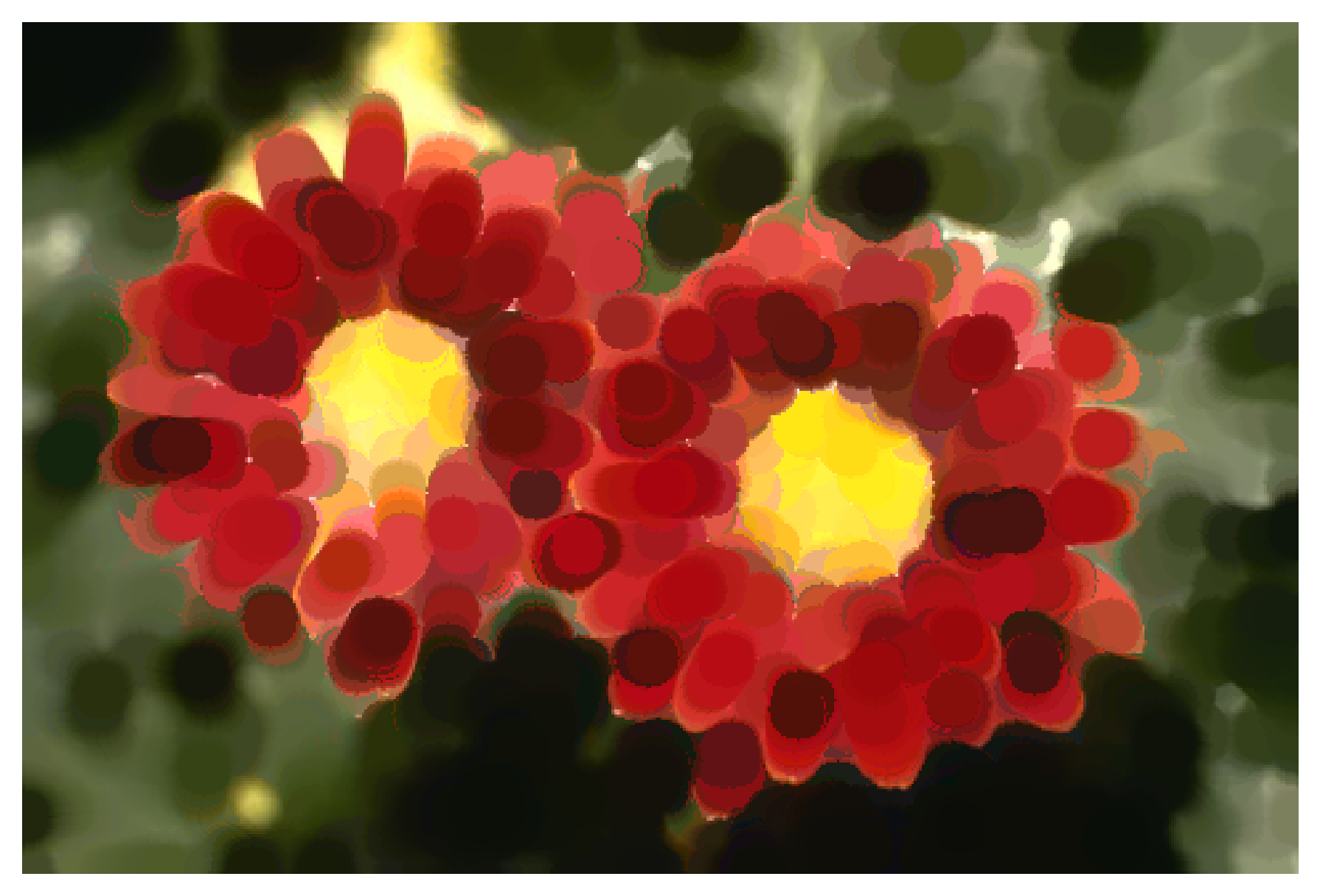} \\
    $\phi_S^{h_1}(\imJ)$ & $\phi_S^{h_2}(\imJ)$ & $\phi_S^{h_3}(\imJ)$\\
    \includegraphics[width=0.32\linewidth]{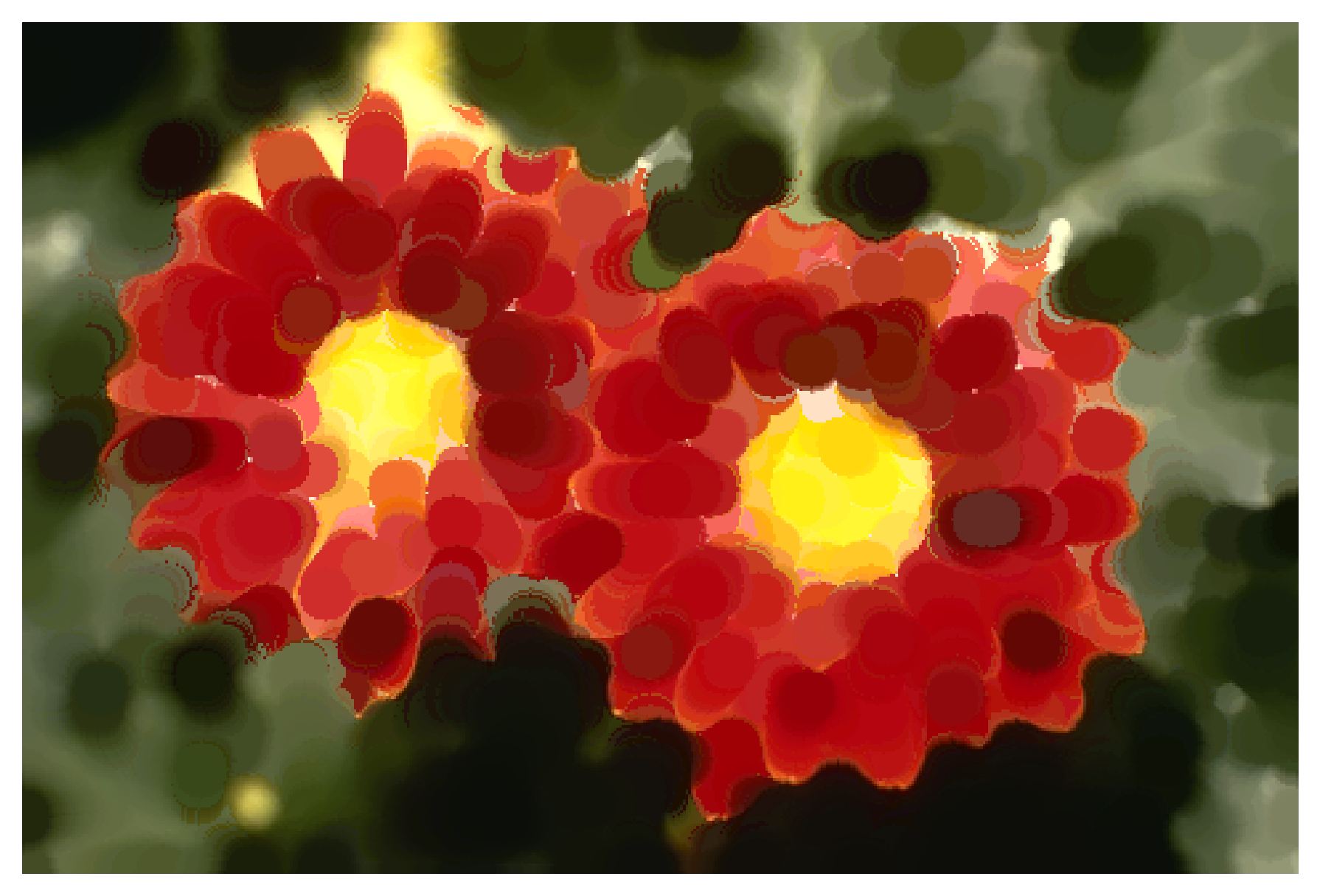} & 
    \includegraphics[width=0.32\linewidth]{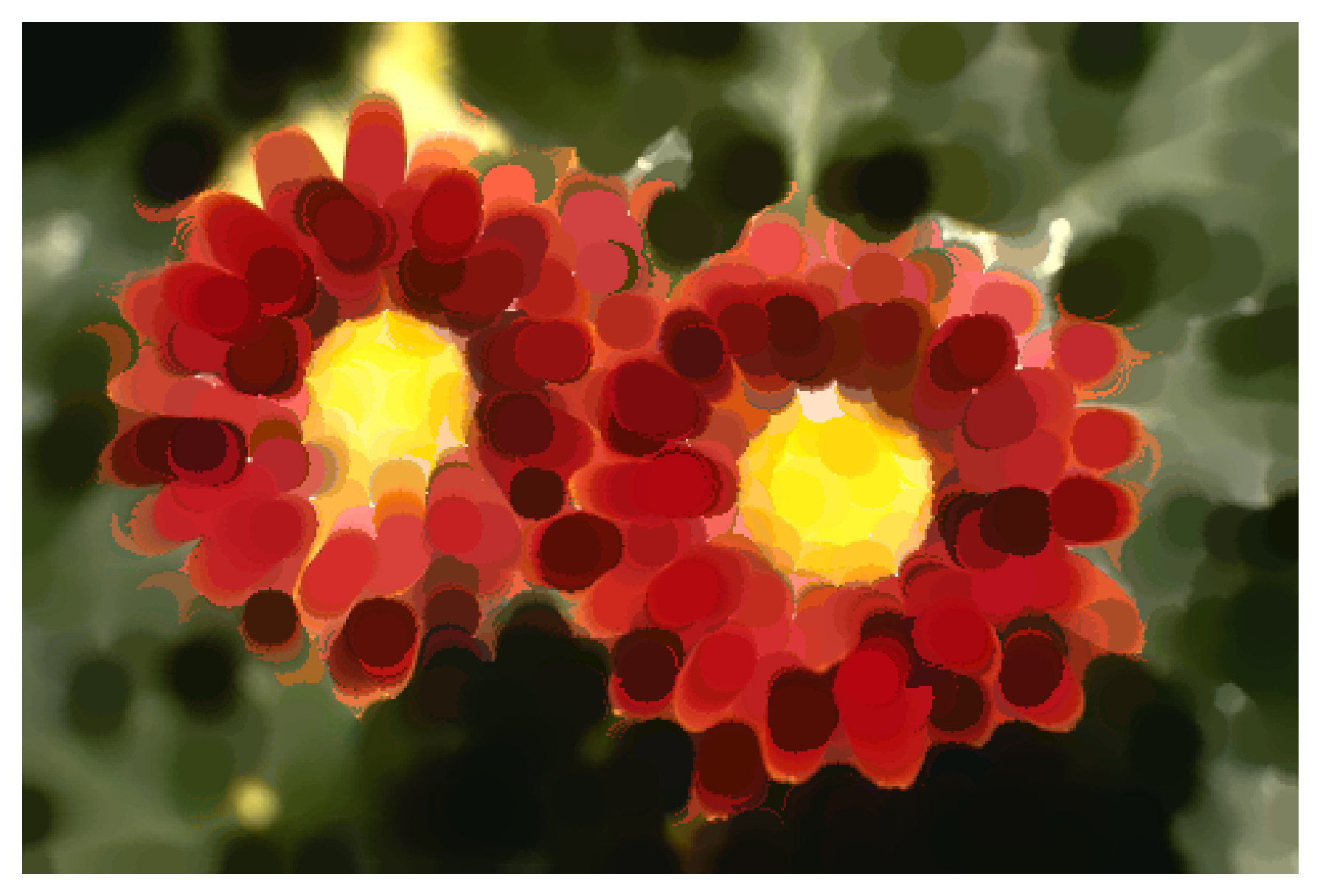} & 
    \includegraphics[width=0.32\linewidth]{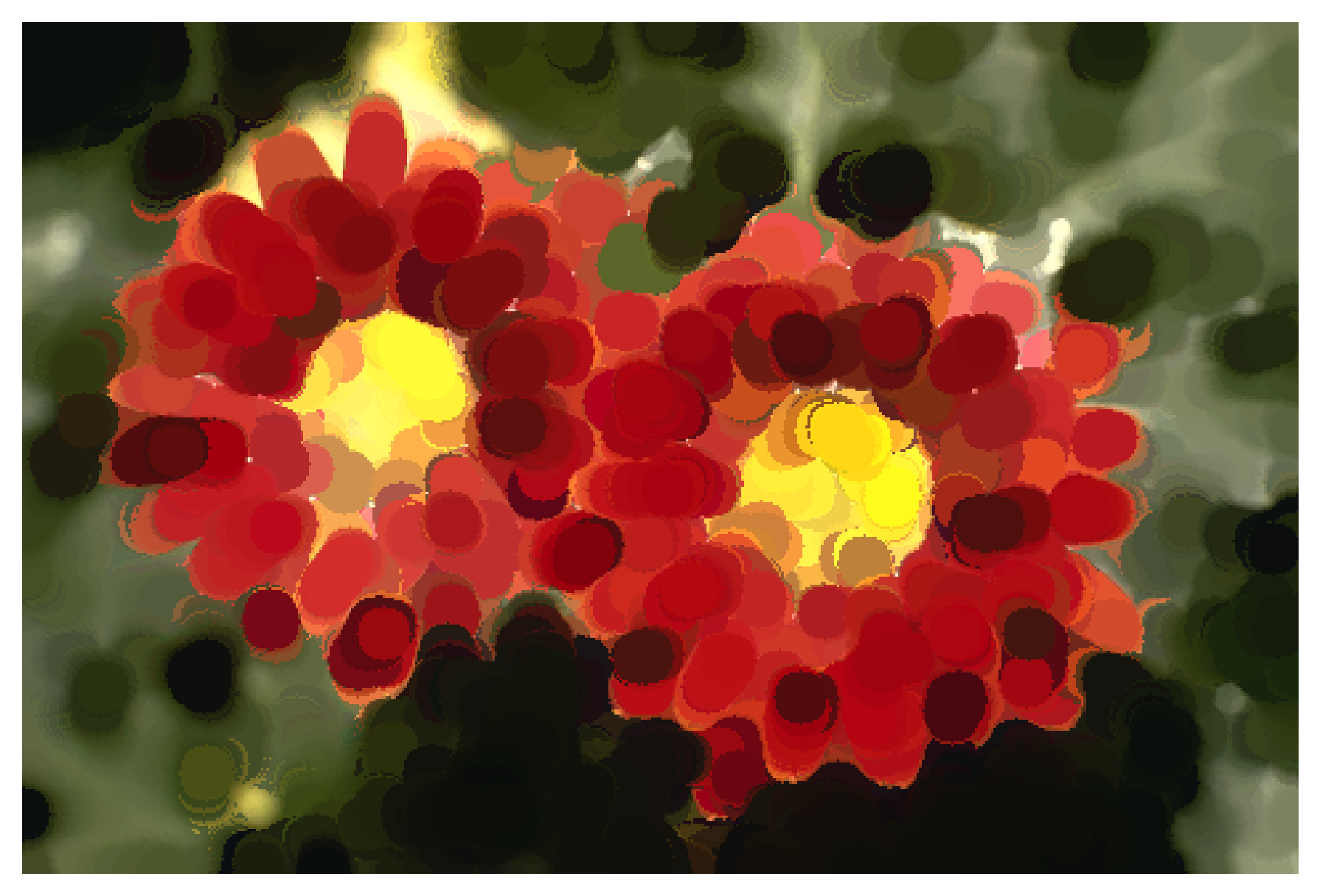}
    \end{tabular}
    \caption{First row: A color image alongside its closing processed using Borda rule and the Condorcet \( h^* \)-mapping. Second row: Closing by Lexicographic orderings  \( h_1 \) ($R-G-B$), \( h_2 \) ($G-B-R$), and \( h_3 \) ($B-R-G$).}
    \label{fig:flowers}
\end{figure}

Although the Condorcet \( h^* \)-mapping was learned using the CIFAR dataset, it can also be applied to process other color images. \new{For instance, the first row in Figure \ref{fig:flowers} displays a color image \( \imJ \) with \( 321 \times 481 \) pixels from the BSD dataset \cite{BerkeleyDataset}, along with the closings $\phi_S^{B}(\imJ)$ and $\phi_S^{h^*}(\imJ)$ obtained using the Borda rule and the Condorcet \( h^* \)-mapping. The second row shows the closings obtained using the three lexicographic orderings. All the openings have been computed using a disk of radius equal to ten pixels.}

\begin{figure}[t]
    \centering
    \begin{tabular}{cc}
    $h_1$: & \fbox{\parbox{0.9\linewidth}{\includegraphics[width=\linewidth]{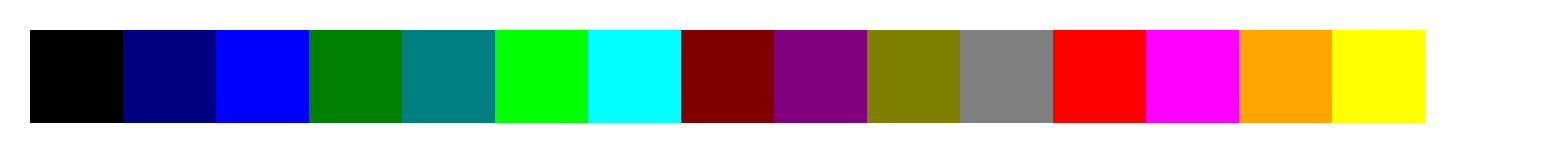}}} \\
    $h_2$: & \fbox{\parbox{0.9\linewidth}{\includegraphics[width=\linewidth]{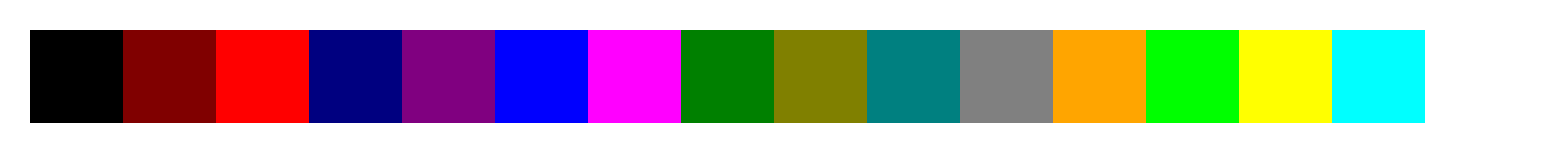}}} \\
    $h_3$: & \fbox{\parbox{0.9\linewidth}{\includegraphics[width=\linewidth]{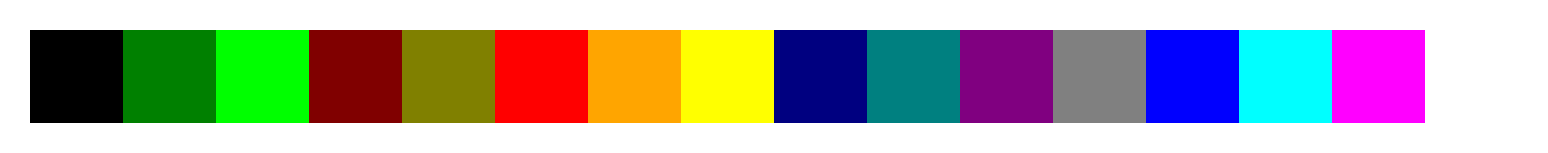}}} \\
    $B$: & \fbox{\parbox{0.9\linewidth}{\includegraphics[width=\linewidth]{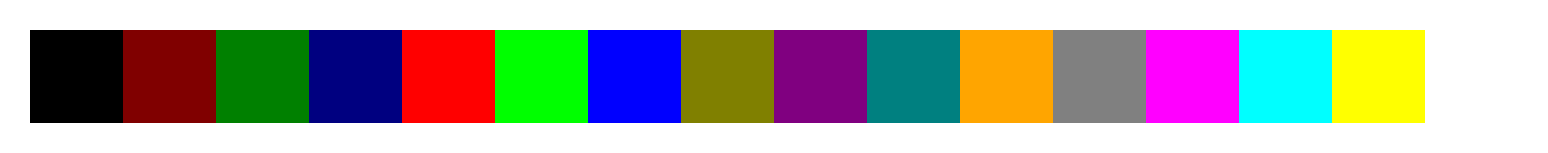}}} \\
    $h^*$: & \fbox{\parbox{0.9\linewidth}{\includegraphics[width=\linewidth]{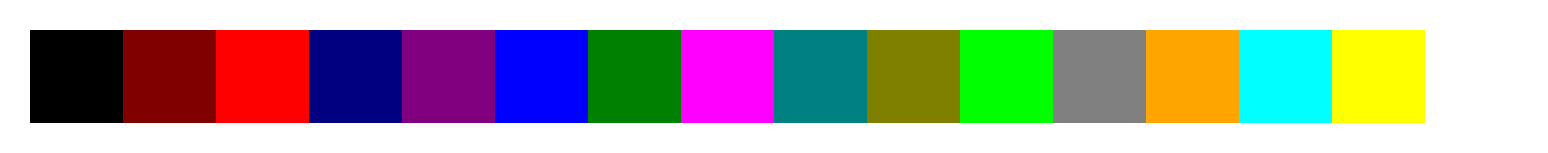}}} \\
    \end{tabular}
    \caption{Illustrative example of the ranking obtained using the lexicographic orderings and the Condorcet \( h^* \)-mapping.}
    \label{fig:colors}
\end{figure}

To conclude this section, Figure \ref{fig:colors} shows the rankings of some colors yielded by the reduced mappings \( h_1 \), \( h_2 \), \( h_3 \), \new{the Borda rule, and the Condorcet \( h^* \)-mapping.} Note that \textit{black} ranks as the least color among the three lexicographical orderings, making it the Condorcet winner (in the sense of ``less than or equal''). Therefore, \textit{black} is also the least color in the Condorcet ordering. When \textit{black} is excluded, there is no consensus among the lexicographical orderings for the remaining colors. In this scenario, the \( h^* \)-mapping yields \textit{maroon}, the second least element in the lexicographical $G-B-R$ ordering, as the second least color in the Condorcet ordering. Dually, \textit{white} is the Condorcet winner (in the sense of ``greater than or equal'') since it ranks as the most preferred color in all three lexicographical orderings. Thus, \textit{white} is also the largest color in the Condorcet ordering. Similar to the previous scenario, there is no consensus on the second-largest color among the three lexicographical orderings. The \( h^* \)-mapping yields \textit{cyan}, the second largest color in the lexicographical $G-B-R$ ordering, as the second largest color in the Condorcet ordering. In general, the ranking of colors yielded by $h^*$ agrees with the Condorcet order derived from the three lexicographic orderings, confirming the proposed approach's potential. 
%%%%%%%%%%%%%%%%%%%%%%%%%%%%%%%%%%%%%%%%%%%%%%%%%%%%%%%%%%%%%%%%%%%%%
%%%%%%%%%%%%%%%%%%%%%%%%%%%%%%%%%%%%%%%%%%%%%%%%%%%%%%%%%%%%%%%%%%%%%

\section{Conclusions and Perspectives}
\label{sec:conclusions}

This paper presents an approach for using gradient descent methods to learn a reduced ordering in the Condorcet sense, where the majority-preferred option is determined through pairwise comparisons. A soft version of the original problem is formulated, enabling the application of learning paradigms from image databases. The experimental results demonstrate that the proposed method is effective on color image datasets, producing reduced orders with a lower irregularity index. Future work will explore extensions of this approach to multi-channel images as hyperspectral images in particular represented after source separation into images valued on the simplex of abundances~\cite{Franchi15}, further expanding its applicability and potential impact to remote sensing and medical applications.

\begin{credits}
\subsubsection{\ackname} This research was jointly funded by FAPESP (grant number 2023/03368-0) and ANR (23-CE23-0028-01) in the Deep Ordering for Vector-Valued Mathematical Morphology and Neural Networks – DEEPORDER project.

% \subsubsection{\discintname}
% It is now necessary to declare any competing interests or to specifically
% state that the authors have no competing interests. Please place the
% statement with a bold run-in heading in small font size beneath the
% (optional) acknowledgments\footnote{If EquinOCS, our proceedings submission
% system, is used, then the disclaimer can be provided directly in the system.},
% for example: The authors have no competing interests to declare that are
% relevant to the content of this article. Or: Author A has received research
% grants from Company W. Author B has received a speaker honorarium from
% Company X and owns stock in Company Y. Author C is a member of committee Z.
\end{credits}


\bibliographystyle{splncs04}
\bibliography{references}

\end{document}